\documentclass{article}

\usepackage{PRIMEarxiv}

\usepackage[utf8]{inputenc} % allow utf-8 input
\usepackage[T1]{fontenc}    % use 8-bit T1 fonts
\usepackage{hyperref}       % hyperlinks
\usepackage{url}            % simple URL typesetting
\usepackage{booktabs}       % professional-quality tables
\usepackage{amsfonts}       % blackboard math symbols
\usepackage{nicefrac}       % compact symbols for 1/2, etc.
\usepackage{microtype}      % microtypography
\usepackage{lipsum}
\usepackage{fancyhdr}       % header
\usepackage{graphicx}       % graphics
\graphicspath{{media/}}     % organize your images and other figures under media/ folder
\usepackage{listings}
\usepackage{dialogue}
\usepackage{mdframed}
\usepackage{xcolor}
\usepackage{subcaption}
\usepackage{amsmath}%
\usepackage{MnSymbol}%
\usepackage{wasysym}%
\usepackage{authblk}
\usepackage{hyperref} 
\graphicspath{{media/}}   
\usepackage{xcolor}
\usepackage{subcaption}
\usepackage{booktabs}
\usepackage{hyperref} 
\usepackage{cleveref}
\usepackage{mdframed}
\usepackage{dialogue}
\usepackage{array}
\usepackage{multirow}
\usepackage{ragged2e}
\usepackage{tabularray}
\usepackage{adjustbox}

\title{Sniff AI: Is My ‘Spicy’ Your ‘Spicy’? Exploring LLM’s Perceptual Alignment with Human Smell Experiences}

\author[1]{Shu Zhong}
\author[2]{Zetao Zhou}
\author[1]{Christopher Dawes}
\author{Giada Brianz}
\author{Marianna Obrist}
\affil[1]{Department of Computer Science, University College London, United Kingdom}
\affil[2]{Division of Psychology and Language Sciences, University College London, United Kingdom}
\renewcommand{\Authfont}{\bfseries}

\begin{document}
\maketitle
\vspace{-40pt}
\begin{figure}[h]
  \includegraphics[width=\linewidth]{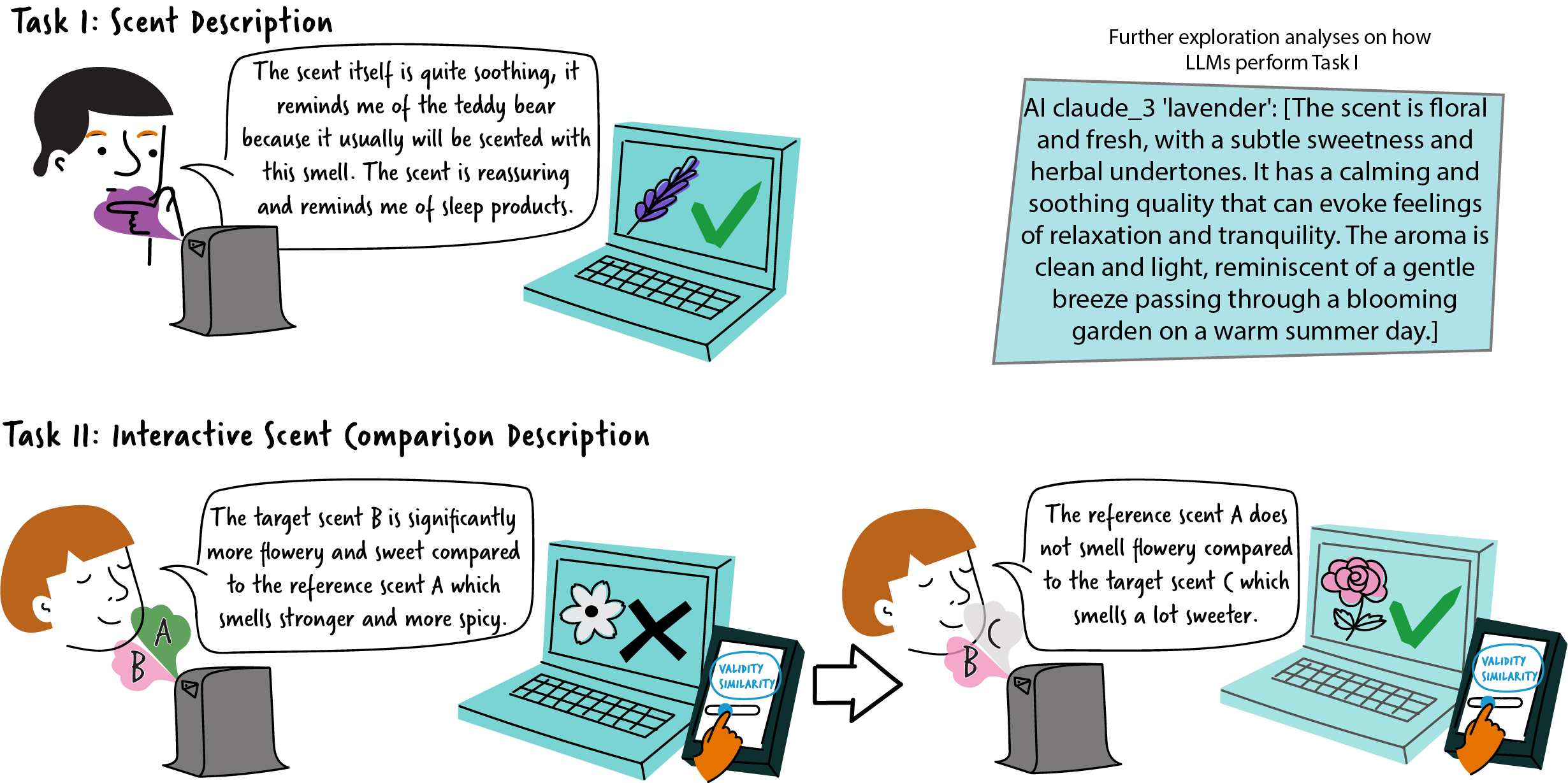}
  \caption{SniffAI: an overview of user study and examples from the user study.}
  % \Description{d.}
  \label{fig:teaser}
\end{figure}

\begin{abstract}
Aligning AI with human intent is important, yet perceptual alignment—how AI interprets what we see, hear, or smell—remains underexplored. This work focuses on olfaction, human smell experiences. We conducted a user study with 40 participants to investigate how well AI can interpret human descriptions of scents. Participants performed "sniff and describe" interactive tasks, with our designed AI system attempting to guess what scent the participants were experiencing based on their descriptions. These tasks evaluated the Large Language Model's (LLMs) contextual understanding and representation of scent relationships within its internal states - high-dimensional embedding space. Both quantitative and qualitative methods were used to evaluate the AI system's performance. Results indicated limited perceptual alignment, with biases towards certain scents, like lemon and peppermint, and continued failing to identify others, like rosemary. We discuss these findings in light of human-AI alignment advancements, highlighting the limitations and opportunities for enhancing HCI systems with multisensory experience integration.
\end{abstract}

\section{Introduction}
Aligning Artificial intelligence (AI) behaviour with human preference is critical for the future of AI. An important yet often overlooked aspect of this alignment is the perceptual alignment. Perceptual alignment refers to the agreement between AI assessments and human subjective judgments across different sensory modalities, such as vision, hearing, taste, touch, and smell \cite{zhong2024exploring, marjieh2023large, sucholutsky2023getting}. It enables AI to better understand the physical world as humans experience it, ensuring that AI applications are reliable and beneficial in real-world settings. For example, consider autonomous vehicles: if the "AI eye" misinterprets data from sensors such as cameras and fails to recognize obstacles or pedestrians, it poses significant safety risks \cite{broggi2016intelligent}. Beyond safety considerations, perceptual alignment plays a critical role in everyday AI applications \cite{sucholutsky2023getting}, whereas olfactory alignment remains completely unexplored. Imagine a future where AI assistants are capable of controlling environmental factors like lighting and scents based on user requests. For instance, instead of asking "Alexa, play uplifting workout music", you can ask Alexa to "spice up my workout session" or "help me remember my holiday to Madrid". Here, the challenge for AI goes beyond playing music and ventures into "AI sniff", to select the ideal scent aligned with human descriptions. This poses the question of how AI would understand and interpret scents in a way that resonates with our personal sensory experiences. Or simply: Is my "spicy" AI's "spicy"?

% The human olfactory system (i.e., sniffing action) detects, decodes, and processes scents, directly sending this information to the brain's limbic system, generating the appropriate internal representations in the brain \cite{maggioni2020smell, cornelio2023smell}. Despite advances in understanding the underlying neurological mechanisms of the sniffing action \cite{bushdid2014humans, snitz2019smellspace} and technological advances in stimulating the sense of smell \cite{cornelio2023smell}, there are still many unanswered questions related to how humans perceive and experience smell. What we, however, know is that smell is uniquely linked to our emotional responses and memories \cite{maggioni2020smell, becsevli2024smell}, making it a compelling modality for enhancing human-computer interaction and experience design, especially in light of AI advancements. 

AI, or mostly Large Language Models (LLMs), lack human-like perceptions; rather, they process human provided language inputs using algorithms that function on binary systems to analyse information \cite{devlin2018bert,liu2019roberta,raffel2020exploring,brown2020language,ouyang2022training}. 
We place a special emphasis on LLMs as they are being increasingly viewed as the interface for human interaction, with agentic AI systems and general alignment being one of its core research domains \cite{shavit2023practices, hendrycks2020aligning}. LLMs, or any neural network-based learning systems, represent concepts and ideas in what is commonly known as an embedding space -- a learned internal high-dimensional vector space \cite{elhage2022toy, nanda2023fact}. In this space, semantically similar items cluster closely together, and the semantic differences between items are preserved \cite{hinton2002stochastic}. This naturally lends us a hand in analyzing how closely aligned humans and AI are. If two items are deemed similar within the AI system's embedding space, humans should also perceive them as similar, if human-AI alignment exists.
Our work then exploits exactly this approach and focuses on analyzing the model's embedding space. We leverage LLM-based embedding models to develop an AI system capable of performing the "Human sniff and describe and AI guesses" task. Here, the LLM encoder translates human language descriptors of scents into the embedding space, and the system makes then scent suggestions based on the semantic similarities measured in this vector space.
% This mode of computation has naturally show sensory preference. Indeed, visual or audio data can be easily digitised as images or recordings. However, this computational bias presents a significant challenge in aligning with more complex human senses such as the sense of smell. 
Given the recent advancements in LLMs' ability to interpret human language, an intriguing question arises: can LLMs effectively understand scents based on user descriptions? For instance, will both LLMs and humans agree on Jasmine and Ylang-Ylang being perceptually similar?

To investigate this question, we conducted an in-person user study with 40 participants, where participants engaged in interactive tasks where an AI system had to guess what scent they were experiencing. 
These participants, who were non-domain experts, were specifically chosen to reflect common perceptions of scents as conceptualized by Henning's Odour Prism \cite{henning1916geruch}, the Fragrance Wheel \cite{edwards2010fragrances}, and attributes such as the fresh, citrusy, and zesty qualities typically associated with lemon. We analysed the system's performance using both quantitative and qualitative methods to capture the participant feedback. Our findings suggest that scent-related semantics are represented in the embedding space, though to a limited extent. There is some degree of perceptual alignment, but it was biased toward certain scents and characteristics. For instance, sometimes the AI believed the human description of \textit{"an aromatic plant that probably you use for like stew or for like chicken and it's very green and is fresh"} refers to eucalyptus rather than rosemary. Occasionally, participants were surprised by certain emergent behaviours; for instance, the AI correctly identified a \textit{"intense masculine scent"} as oakmoss. We discuss these findings in light of recent advancements of LLMs and efforts towards improved human-AI alignment, with the opportunity to enhance HCI systems with multisensory experience integration.

% Specifically, we assess how accurately LLMs can predict the odor a human is experiencing based on descriptions of olfactory descriptors. Using these descriptors, LLMs attempt to identify target scents by analyzing similarities within their high-dimensional embedding spaces.

% emergent behaviour

\section{Background and Related Work}
\subsection{Human-AI Interaction and Perceptual Alignment}
Human-AI alignment involves designing, developing, and refining AI systems to understand, predict, and enhance human intentions and actions, whilst generating informative, harmless, and helpful responses \cite{hendrycks2020aligning, ziegler2019fine, sucholutsky2023getting}. A notable contribution is from Hendrycks et al. \cite{hendrycks2020aligning}, who introduced the ETHICS dataset to evaluate language models' understanding of fundamental moral concepts, such as justice and well-being. Recent studies emphasize integrating human-in-the-loop to reduce toxicity and foster ethical behaviour \cite{ouyang2022training, goyal2022your, pan2023rewards}. 

There is growing research into representational alignment between AI and humans. Representation alignment refers to the extent to which the internal representations of two or more information processing systems are aligned \cite{sucholutsky2023getting}. Peter et al. \cite{peter2023we} compared the performance of general-purpose embedding models, such as OpenAI's \texttt{text-embedding-ada-002}, with domain-specific models, such as the CLAP text-audio embedding model\cite{elizalde2023clap} in capturing nuances in expressive piano performances. In this particular case, all models showed a degree of alignment, and general-purpose models outperformed domain-specific ones. 

Extending this exploration of Human-AI alignment into sensory judgments, Lee et al. \cite{lee2023visalign} introduced the VisAlign dataset, designed to evaluate the alignment between AI and human visual perception. This dataset aids in understanding how to better align AI with human vision perceptual processes, particularly in vision. Another exemplified research is done by Zhong et al. \cite{zhong2024feeling}, who investigated the agreement between humans and six vision-based Multimodal Large Language Models (MLLMs) in describing tactile qualities of textiles, they found these models' descriptions were detached from human descriptions in both sentiment and word usage. Differently, Marjieh et al. \cite{marjieh2023large} demonstrated that GPT-4 can effectively interpret certain human sensory judgments (e.g., colour, sound and taste) based on textual sensory inputs. For example, they presented the same pair of colours (red and blue) to both humans and GPT models\footnote{Given that GPT models lack the ability to "see" colour hex codes are provided as textual inputs.}, requesting each to provide a similarity rating and then comparing the resulting scores. Their findings showed that GPT models produced judgments that correlated with those of humans. To the best of our knowledge, the alignment between AI and human olfaction remains unexplored. Our research aims to extend these foundational studies by integrating semantic embeddings \cite{bengio2013representation} to investigate olfactory perception.

AI systems process information differently from humans, by representing concepts and ideas within a latent space or embedding space \cite{elhage2022toy, nanda2023fact}. These spaces allow the AI to efficiently encode, process, and manipulate complex information, capturing relationships and patterns in the data \cite{hoff2002latent, hinton2002stochastic}. In embedding spaces, embeddings are learned representations, similar items are grouped closely together, while differences are preserved through distance and direction \cite{hinton2002stochastic, bricken2023towards}. The embedding model is a fundamental component of the LLM architecture, and studying the encoder helps explore how the model interprets and organizes information at its core - both dimensional and categorical structure \cite{bricken2023towards}. Understanding the encoder's function is crucial, as it provides insights into how the LLM structures its internal knowledge \cite{elhage2022toy}. Zhong et al. \cite{zhong2024exploring} explored textile tactile experiences using an LLM embedding model, finding limited perception alignment and biases toward specific textiles.
In this study, we investigate the ability of LLMs to understand human smell experiences by analysing how they represent scents within their embedding models. By examining how these models encode olfactory concepts, we aim to understand how they capture both the similarities and the relationships between different scents. This understanding of LLMs' internal processes can offer a more meaningful grasp of the internal workings of LLMs and potentially enhance their performance \cite{elhage2022toy, mikolov2013efficient}.

\subsection{Growing Importance of Smell in HCI and Experience Design}
The integration of olfactory experiences in HCI has been a relatively underused sense compared to vision and sound technologies \cite{olofsson2017beyond, brooks2023third}. However, olfactory experiences hold the potential to significantly enhance virtual reality environments and add a new dimension to HCI applications, particularly by facilitating attention shifts  \cite{keller2011attention}. In traditional audio-visual interactive media, scent is frequently used as a supplementary tool to deepen immersion \cite{carter2023scent, becsevli2024smell}. From an anthropological perspective, olfaction is deeply intertwined with human culture and may serve as a primary sensory modality to enrich multisensory experiences \cite{classen1997foundations}, especially in digital applications that emphasise voluntary engagements, such as memory recall, attention, spatial orientation, and wellbeing \cite{olofsson2017beyond, becsevli2023nose, maggioni2020smell, dmitrenko2017ospace}. Yet, many questions about human perception and experience of smell remain unanswered. What we, however, know is that smell is uniquely linked to our emotional responses and memories \cite{maggioni2020smell, becsevli2024smell}, making it a compelling modality for enhancing HCI and experience design, especially in light of AI advancements.

There is a growing interest within the HCI community in integrating olfactory devices into everyday life \cite{carter2023scent}. For example, Brooks et al. \cite{brooks2023smell} introduced the "Smell \& Paste" toolkit, a scratch-and-sniff prototyping tool that allows novices to quickly create personalized olfactory experiences. Additionally, toolkits like OWidgets have been developed to enable the creation and replication of olfactory experiences in HCI, beyond the traditional audio-visual design space \cite{MAGGIONI2019248}. 
% We aimed to explore the integration of smell experiences into daily activities through natural interactions. Therefore, we recruited the general public rather than domain experts for our AI alignment task. 

% [show growing efforts towards integration smell, applications as you have in section 2.3., and some of the basic insights we have in designing interactions and experiences with smell... drawing on some parts in section 2.1 and 2.3, what is missing is the common language]

\subsection{Bridging Human and Machine Understanding of Smell Experiences}\label{sec:rl:hci}
The olfactory sense presents a unique challenge for cognitive science research, particularly in the realms of perception, memory, and language \cite{iatropoulos2018language, majid2021human,obrist2014opportunities}.  
To understand olfaction, researchers employ a variety of behavioural, neurophysiological, computational, and theoretical methods to study how smells are perceived and represented in the brain and language \cite{mori1994emerging, theimer2012fragrance,bushdid2014humans, snitz2019smellspace,brookes2011olfaction, iatropoulos2018language}. From a human perspective, conveying smell experiences is more challenging than describing colours, which benefit from a standardized vocabulary (e.g., adjectives like "red" and hex codes). Scents lack an equivalent standardized language\cite{schifferstein2005capturing}, making it challenging for the general public, without specialised knowledge, to identify scents based solely on chemical names \cite{wilson2006learning}. To address this, classification systems such as Henning's Odour Prism \cite{henning1916geruch} and the Fragrance Wheel \cite{edwards2010fragrances} organize scents based on their olfactory characteristics. 
% In our study, we employed the Fragrance Wheel's four main families: Fresh, Floral, Oriental, and Woody, to select our scents.

Building on the challenges of scent categorization, recent linguistic and AI research has explored the connection between molecular structures and odour perception or multimodal representation \cite{lee2023principal, keller2017predicting, lisena2022capturing, oderopa2020smell}. Studies like the DREAM Olfaction Prediction Challenge by Keller et al. \cite{keller2017predicting} have aimed to understand how humans perceive different molecules as scents. Similarly, the Odeuropa project \cite{oderopa2020smell} has significantly contributed to digital heritage by creating a smell-linguistic odour dataset \cite{lisena2022capturing} and launching a multimodal data challenge that integrates vision and text to categorize sniff behaviours in digital heritage \cite{zinnen2022odor}. Lee et al. \cite{lee2023principal} used a Graph Neural Network (GNN) to predict olfactory descriptors from molecular structures, showing AI's growing capability in olfaction. Additionally, mainstream AI in this field often utilizes electronic noses (e-nose) with Interdigitated Electrode (IDE) structures and Molecular Imprinted Polymer (MIP) sensors for detecting specific chemicals such as limonene \cite{hawari2012recognition}. These developments indicate that with sufficient data, AI models have the potential to match or even exceed human capabilities in olfactory perception, whereas the smell experience in existing state-of-art-art models, like LLMs, remains underexplored.

Previous studies have used word embeddings to learn sensory description languages in natural language processing \cite{alvarado2023organization}. However, word embeddings often fail to capture the nuanced sentiment and contextual meanings of entire texts, as they represent words as static vectors without considering variations in usage across different contexts \cite{mikolov2013efficient}. In addition, the nature of scent is multidimensional, extending beyond a single descriptor. To address these complexities, our work adopts a more sophisticated approach by using sentence-level embeddings generated from LLM encoders. LLM encoders process entire sentences rather than individual words, allowing them to capture the semantic, rather than focusing solely on lexical or stylistic elements. This results in context-aware embeddings that adapt to the surrounding text, providing a richer and more comprehensive analysis of the content's sentiment and meaning \cite{devlin2018bert,brown2020language}.

\section{The Sniff AI System Design}\label{sec:method}

\begin{figure}[t]
  \centering
  \includegraphics[width=\linewidth]{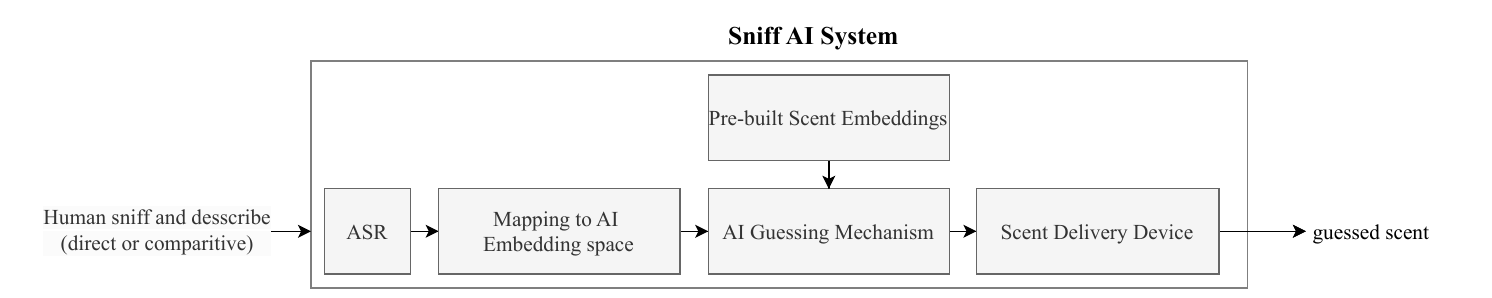}
\caption{An overview of Sniff AI System components and workflow: starting with a human describing a scent. The description is processed by an ASR system, mapped to an AI embedding space, and compared with pre-built scent embeddings. The AI guessing mechanism predicts the scent and then delivers the guessed scent by a scent delivery device.}
% \Description{The diagram shows the workflow of the "Sniff AI System," starting with a human describing a scent. The description is processed by an ASR (Automatic Speech Recognition) system, mapped to an AI embedding space, and compared with pre-built scent embeddings. The AI guessing mechanism predicts the scent, which is then delivered by a scent delivery device as the guessed scent output.}
\label{fig:systemoverview} 
\end{figure}

This section introduces Sniff AI, our system to investigate whether LLMs can effectively understand smells based on human descriptions. We would like to understand how AI interprets human-described scent differences by analyzing the vector relationships in LLM's latent embedding space. We begin by discussing how LLMs encode concepts like scent descriptions into high-dimensional vector spaces (embeddings) and provide an overview of the Sniff AI system design in \Cref{sec:method:repre}. Following this, we then detail the design of our Sniff AI system with five core components as shown in \Cref{fig:systemoverview}: Automatic Speech Recognition (ASR) and pre-built scent embedding generation (described in \Cref{sec:method:rep_scent}), mapping human smell experiences into the AI's embedding space (\Cref{sec:method:map}), the AI guessing mechanism (\Cref{sec:method:prediction}), and scent delivery (\Cref{sec:method:scentselect}). We also discuss how we selected the scents in \Cref{sec:method:scentselect}.

% We evaluate the AI's understanding of scents through two tasks embedded within the scent-guessing system.

\subsection{Scent Embedding Space and the Sniff AI System Overview
}\label{sec:method:repre}
AI systems "think" differently from humans, as they represent concepts and ideas within an embedding space \cite{elhage2022toy, nanda2023fact}. 
% Embeddings are learned representations; items with similar meanings are grouped closely together, while differences are preserved by distance and direction \cite{hinton2002stochastic, bricken2023towards}. 
While the decoder-only models like GPT-4 \cite{achiam2023gpt} and Gemini \cite{team2023gemini} are for generating output in an auto-regressive fashion, the encoder-based models, or embedding models, are fundamental for understanding how the model interprets and organizes information \cite{neelakantan2022text, bricken2023towards}. 
For example, in an LLM embedding space, the distance between "man" and "woman" is comparable to that between "king" and "queen", reflecting their analogous relationships. The direction between vectors from the model's embedding space can also reflect semantic differences and similarities, such as the shift from "man" to "woman" mirroring the shift from "king" to "queen", both indicating a change in gender \cite{sanderson2024neural}. 
By studying how LLMs encode human scent descriptions, we gain insight into how they capture both similarities and the nature of relationships between scents. 
% Specifically, we use state-of-the-art LLM embedding model--OpenAI's \texttt{text-embedding-3-large} model \cite{neelakantan2022text}.

\subsubsection{Scent Embedding Space}
In a scent embedding space, if AI's perception aligns with human experience, similar scents should be located near each other (in terms of their vector norms, such as $l_p$ norms), while dissimilar ones should be positioned further apart, with the direction and distance between them representing their differences. For example, "lemon" and "lime" might be near each other due to their similar citrus characteristics, while "rose" and "jasmine" could be close as they are both floral. The placement, distance, and direction of these scent vectors, obtained from the LLM embedding model, should reflect meaningful olfactory relationships. If a scent is described as sweeter than "lemon," we would expect AI to point out "vanilla" is closer rather than "peppermint", suggesting a transition from a fresh to a sweet scent. Conversely, an embedding vector from "sandalwood" to "peppermint" might indicate a transition from a warm, woody scent to a cool, minty one. These vector mappings are key to understanding how AI interprets and differentiates between various scents.

\subsubsection{An overview of the "Sniff AI" study and system}
We conducted a user study, where we use the Sniff AI system to identify scents based on human descriptions. The Sniff AI system recognizes human voice input, predicts, and delivers scents in real-time. The study's objective is to evaluate the alignment between human sensory descriptions and the AI's scent judgement via two tasks: the Scent Description Task and the Interactive Scent Comparison Task. We describe further these tasks in \Cref{sec:method:map} and the detailed user study in \Cref{sec:study}. 
% Both tasks assessed how well the AI could understand and interpret human olfactory experiences. Task 2, in particular, focused more on the alignment between human and AI olfactory perception in their representations.
% The Sniff AI system recognizes human voice input, predicts, and delivers scents in real-time, as shown in \Cref{fig:systemoverview}. Initially, we constructed a scent knowledge base using pre-built scent embeddings, detailed in \Cref{sec:method:rep_scent}. To map human verbal descriptions, we utilized ASR to convert voice into text, which is then encoded into the embedding space (\Cref{sec:method:map}). We describe the AI's guessing mechanism in \Cref{sec:method:prediction}, where it predicts the scent that most closely matches based on semantic similarity. And last, the predicted scent is delivered to human.

\subsection{Pre-encoded Scent Embeddings}
\label{sec:method:rep_scent}

To evaluate the AI's ability to understand and distinguish between scents, we first need to generate scent representations in a form suitable for AI processing, we call this process pre-encode scent embeddings. This includes converting scent-related data into numerical vectors within the LLM's high-dimensional embedding space. 

We consulted and worked with domain experts to create catalogue descriptions for each scent. These catalogue descriptions provide essential information about each scent's source and composition. For example, \textit{"The scent of Rosemary is from its essential oil. The essential oil of Rosemary is extracted from the Rosmarinus Officinalis plant (CAS \footnote{Chemical Abstracts Service (CAS) registry number, a unique numerical identifier for the chemical substance.}: 8000-25-7)"}. 
These catalogue descriptions (${x}_{i}$) were encoded into embedding vectors using OpenAI's \texttt{text-embedding-3-large} model with 3072 dimensions \cite{neelakantan2022text}, defined as:

\begin{equation}
    v_i = f_{encoder}(x_i)
    \label{equ:encode}
\end{equation}

A total of 20 unique vectors were generated since we have 20 scents, each representing a different scent (${v}_{i}$). These vectors, known as $\mathcal{E}_{scent} = \{v_1, v_2, \ldots, v_{20}\}$, form the basis of the AI's scent knowledge and are used repeatedly throughout this study, as referenced in \Cref{tab:scent}. Detailed information on the scents, including their concentrations, manufacturers, and catalogue descriptions, is provided in the supplementary materials.

\subsection{ASR and Mapping Human Olfactory Experiences to the AI Embedding Space}
\label{sec:method:map}

\begin{figure}[t]
  \centering
  \includegraphics[width=\linewidth]{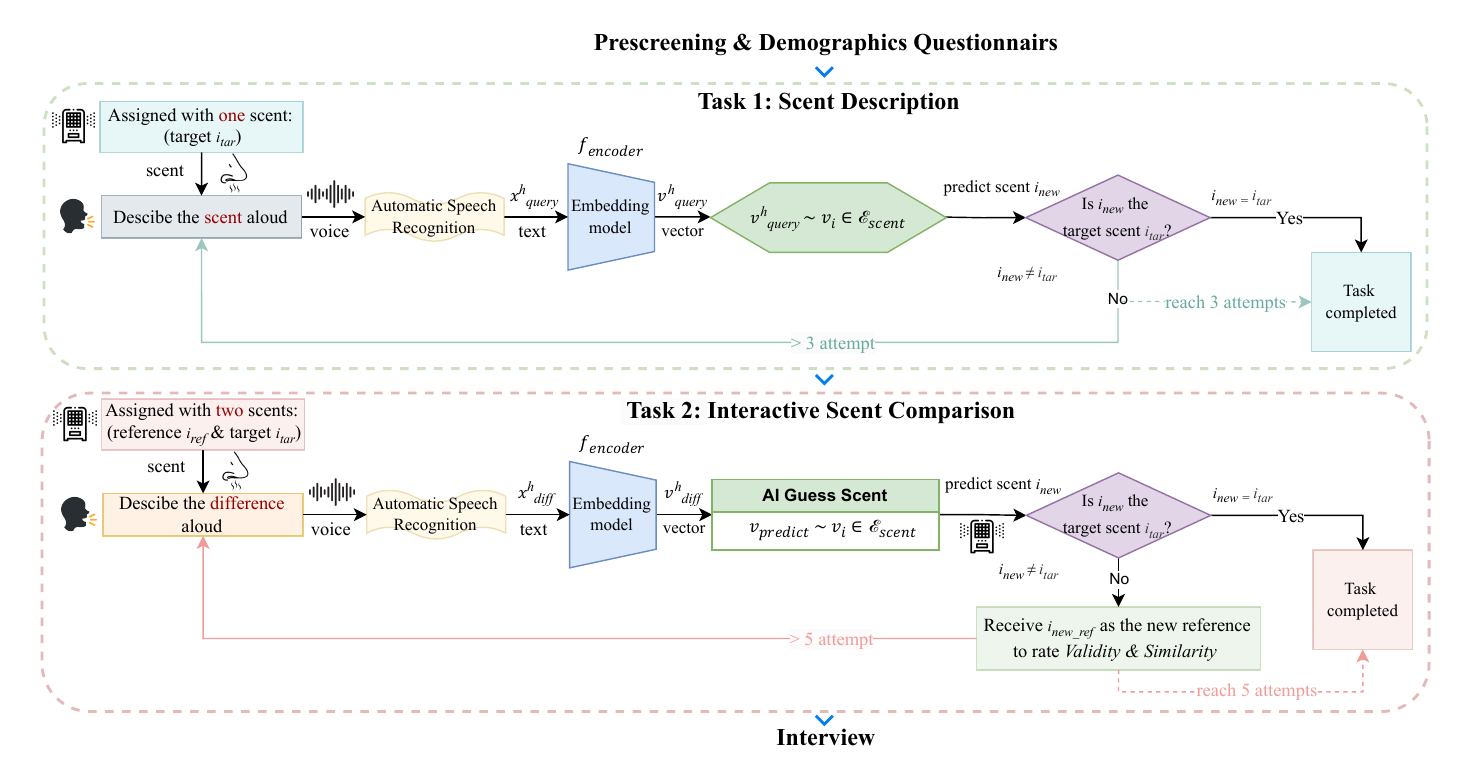}
\caption{An overview of the "AI sniff" method and the "Guess what scent" user study workflow. }
% \Description{An overview of the "AI sniff" method and the "Guess what scent" user study workflow. There are two tasks detailed in each step as out lined in section 3.3 and 3.4}
\label{fig:flow} 
\end{figure}

To explore LLM's embedding space representations, we designed two interactive tasks involving human participants. In both tasks, participants sniff scents and provide descriptions of their smell experiences, which is then encoded by the LLM encoder and serve as representations of LLM's olfactory perception, as illustrated in \Cref{fig:teaser} and detailing the specifics of the implementation  \Cref{fig:flow}: 
\begin{enumerate}
    \item The Scent Description Task: This task (Task 1 in \Cref{fig:flow}) tests the LLMs encoder's ability to match a specific scent based on human-provided descriptions in the latent embedding space. The AI identifies the closest match within its scent embedding space by matching the internal semantic similarity of scent representations.
    \item The Interactive Scent Comparison Task: This task (Task 2 in \Cref{fig:flow}) evaluates the LLMs encoder's ability to understand and represent the transitions between different scents. This task examines whether the AI can reflect the progression from one scent to another using comparative descriptions provided by humans. For example, we examine if the vector from "the scent of mint" to "the scent of rose" in the LLM's embedding space reflects a shift from a fresh scent to a more floral one and aligns with human smell experience.
\end{enumerate}

% Further details on the AI system for scent prediction are presented in \Cref{sec:method:prediction}, our evaluation method in \Cref{sec:study:eval} and the study procedure is presented in \Cref{sec:study:procedure}.
The mapping methods and AI prediction mechanisms associated with these two distinct tasks (Task 1 and 2) then differ slightly.
% We use human smell experiences as representations of human olfactory perception. This section explains how we map these human smell experiences to the AI's embedding space for two distinct tasks. As depicted in \Cref{fig:method}, participants were exposed to various scents by a novel scent delivery device described in \Cref{sec:method:scentselect}. 
Each participant verbally describes their smell experiences. In Task 1, participants describe the smell of a single scent, while in Task 2, they describe the differences between two different scents. These verbal descriptions were captured and transcribed into text using ASR -- OpenAI's \texttt{whisper-1} model \cite{radford2023robust}. 
The transcribed texts were then encoded into numerical vectors using the same encoder, $f_{encoder}$, OpenAI's \texttt{text-embedding-3-large} model \cite{neelakantan2022text}.

\subsubsection{Task 1: Scent Description}\label{sec:method:map1}

Participants provided descriptions for a single scent, which were transcribed as $x^{h}_{single}$ and processed by $f_{encoder}$, resulting in a query vector $v^{h}_{query}$, calculated as follows:

\begin{equation}
    v^{h}_{query} = f_{encoder}(x^{h}_{query})
    \label{equ:h_1}
\end{equation}

\subsubsection{Task 2: Interactive Scent Comparison}\label{sec:method:map2}
Participants described differences between a target scent (\(i_{tar}\)) and a reference scent
(\(i_{ref}\)). The AI system is initially given the embedding of the reference scent, $v_{ref} \in \mathcal{E}_{scent}$, as the starting point for identifying the target scent. The AI uses an updated $v^{h}_{update}$ to make predictions, following a multi-step process. Firstly, the descriptive differences $x^{h}_{diff}$ were encoded into a vector $v^{h}_{diff}$ using Equation \ref{equ:h_diff}. Then, the embedding vector of the reference scent, $v_{ref}$, was combined with $v^{h}_{diff}$. The resultant vector was normalized to obtain the updated query vector $v^{h}_{update}$:

\noindent
\begin{minipage}{.5\linewidth}
\begin{equation}
    v^{h}_{diff} = f_{encoder}(x^{h}_{diff})
    \label{equ:h_diff}
\end{equation}
\end{minipage}%
\begin{minipage}{.5\linewidth}
\begin{equation}
    v^{h}_{update} = \frac{v_{ref} + v^{h}_{diff}}{\| v_{ref} + v^{h}_{diff} \|}
    \label{equ:h_new}
\end{equation}
\end{minipage}

Here, \(\|v\|\) is the Euclidean norm of \(v\), calculated as \(\sqrt{v v^T}\).

\subsection{The AI Guessing Mechanism}
\label{sec:method:prediction}

The general principle of relying on the AI's embedding to make a guess in both Task 1 and 2 are alike.
The AI guessing mechanism performs an information retrieval using the cosine similarity measure defined as:
\begin{equation}
    v_{query} = \arg \max_{v_i \in \mathcal{E}_{scent}} \frac{v_{query} \cdot v_i}{\|v_{query}\| \|v_i\|}
    \label{equ:cos}
\end{equation}

For each prediction, the AI aims to identify a target scent \(i_{tar}\) within $\mathcal{E}_{scent}$. The system selects the most similar embedding from the set of 20 possible embedding vectors $\mathcal{E}_{scent} = \{v_1, v_2, \ldots, v_{20}\}$:

\begin{equation}
    \text{ID} = \text{max}(\text{Cos}(v_{query}, v_i) | v_i \in \mathcal{E}_{scent})
    \label{equ:topk}
\end{equation}

where \(\text{ID}\) represent the indices of the largest similar embedding to the query, $Cos$ computes the cosine similarities between vectors.

\subsubsection{Task 1: Scent Description}
Here, $v^{h}_{query}$ from \Cref{sec:method:map} is used as \(v_{query}\) in \Cref{equ:cos} to identify the scent embedding that most closely matches the description. The system then provides feedback by audibly announcing and visually displaying the guessed scent ID on the interface.

To initiate the task, a target scent \(i_{tar}\) is allocated and diffused to the participant via the novel scent-delivery device, activated by pressing the "sniff the scent" button in our UI. Participants are instructed to describe their smell experiences verbally aloud to the system. The AI then processed the descriptions and made an informed guess, presenting the guessed ID both audibly and on the display interface. If the AI successfully identifies the scent, that round is completed. If not, participants can refine their descriptions, prompting the AI to make another guess based on the new description. This iterative process continues until the scent is correctly identified or a predefined limit is reached (see \Cref{sec:study:eval}).
% This process could be repeated up to three times per round.

\subsubsection{Task 2: Interactive Scent Comparison} In Task 2, participants are involved in an interactive scent comparison. They provide multiple comparative descriptions of the target scent, $x^{h}_{diff}$, defined in Section 3.4. The AI predicts the position of the target scent in the embedding space, $\mathcal{E}_{scent}$, using an updated vector $v^{h}_{update}$. This vector is generated by combining the embedding of the reference scent with the human-provided comparative description (\Cref{equ:h_new}). A cumulative description is learned. The system then delivers the guessed scent via the scent-delivery device.

Each round starts with participants receiving an initial pair of scents: a reference scent (\(i_{ref}\)) and a target scent (\(i_{tar}\)). And AI aims to identify (\(i_{tar}\)). Participants verbally describe the differences between the reference and target scents aloud. These descriptions are captured via ASR in our system and processed by the system to generate a comparative description vector (\(v^{h}_{diff}\)), as described in \Cref{sec:method:map}. The updated query vector \(v^{h}_{update}\) in the later rounds is calculated by integrating \(v_{ref}\) with \(v^{h}_{diff}\), as defined in Equation \ref{equ:h_new}.

\subsection{Scent Delivery and Selection}
\label{sec:method:scentselect}
\begin{table}
\centering
\caption{20 Selection Scent using the Fragrance Wheel \cite{edwards2010fragrances} falling into Four Family and Subfamilies.}
\label{tab:scent}
\begin{tabular}{llll|llll}
\toprule
Families     & Subfamilies     & Scent      & ID & Families & Subfamilies     & Scent        & ID  \\ 
\midrule
             & Aromatic        & Rosemary   & 1  &          & Soft  Oriental  & Vanilla      & 11  \\
             & Aromatic        & Eucalyptus & 2  &          & Soft  Oriental  & Cardamom     & 12  \\
~ ~Fresh~ ~~ & Citrus          & Bergamot   & 3  & Oriental & Oriental        & Frankincense & 13  \\
             & Fruity          & Lemon      & 4  &          & Woody  Oriental & Sandalwood   & 14  \\
             & Green           & Peppermint & 5  &          & Woody  Oriental & Patchouli    & 15  \\ 
\midrule
             & Floral          & Lavender   & 6  &          & Woods           & Pine         & 16  \\
~            & Floral          & Gardenia   & 7  & ~        & Woods           & Cedarwood    & 17  \\
~Floral      & Soft Floral     & Rose       & 8  & ~Woody   & Mossy Woods     & Oakmoss      & 18  \\
~            & Floral Oriental & Jasmin     & 9  & ~        & Dry Woods       & Black pepper & 19  \\
~            & Floral Oriental & Geranium   & 10 & ~        & Dry Woods       & Cinnamon     & 20 \\
\bottomrule
\end{tabular}
\end{table}

The scents were delivered using a digital scent delivery device developed by OW Smell Made Digital \footnote{OW Smell Made Digital: \url{https://www.ow-smelldigital.com/}.}. The device pumps air through individually separated scent channels. We added 250 microlitres of each scent to a cellulose sponge (25mm x 10mm x 1mm) placed into the device. The level of fragrance was determined based on previous studies using the same device \cite{becsevli2023nose}.

We investigate smell experiences using the Fragrance Wheel \cite{edwards2010fragrances} -- a well-established scents classification system based on their scent profile characteristics. For our study, we selected 20 scents, shown in \Cref{tab:scent}, representing a diverse range of profiles by choosing five scents from each of the four Fragrance Wheel families: Floral, Fresh, Woody, and Oriental \cite{edwards2010fragrances}. Careful consideration was given to ensure a balanced variety of scents within each category to avoid making the AI's task too simple, which could lead to ceiling effects. For example, if participants describe the only floral scent as "floral", the AI would immediately guess correctly. However, by having multiple but distinct floral scents, such as rose, geranium, and lavender, the task becomes more challenging. On the other hand, we avoided focusing exclusively on a single scent family (e.g., only florals) to prevent the task from becoming too difficult and to ensure the implications for human-AI alignment extend beyond just one scent family. We cross-reference our selection with the literature presented in Section 2.3 on the Fragrance Wheel \cite{edwards2010fragrances}. To maintain authenticity, all scents used were natural essential oils and extracts to ensure the fragrance closely matched its label (e.g., the lemon scent was derived from lemon peel oil extract). Additionally, we sourced fragrances from the same supplier whenever possible to maintain consistency across scent quality and formulation.

% [ref ScentHaptics]. 

\section{User Study}\label{sec:study}

The user study, as described before, utilized the Sniff AI system to identify scents based on human descriptions, where participants sniffed and provided detailed descriptions of the scents. Our aim was to evaluate the alignment between human sensory descriptions and the AI's scent judgements through two tasks: the Scent Description Task (Task 1) and the Interactive Scent Comparison Task (Task 2). Both tasks were designed to assess how well the AI could understand and interpret human olfactory experiences. 
Task 2, in particular, focused more specifically on evaluating the alignment between human and AI olfactory perception in their respective representations, as discussed in \Cref{sec:method}. The study employed a mixed-methods approach using a repeated-measures, within-subjects design to ensure a robust evaluation of AI's ability to capture human scent perception.

\subsection{Participants}
We aimed to explore the integration of smell experiences into daily activities through natural interactions. Therefore, we recruited the general public rather than domain experts for our AI alignment task. A total of 40 participants (22 female, 18 male; aged 19-50, mean = 28.95, SD = 5.99) were recruited for an in-person user study, targeting the general public rather than domain experts, as detailed in \Cref{sec:rl:hci}. None of the participants had any olfactory sensory impairments that could affect their olfactory perception. The participants came from 14 countries across four continents: Europe (21), Asia (17), North America (1), and Oceania (1), and all were native or highly proficient English speakers. Their professions were diverse, including computer scientists (6), IT and engineers (6), HCI and psychology students (4), bio/medical students (4), neuroscience researchers (3), local government workers(2), a psychotherapist and etc. Participants provided written informed consent before participating in the 60-minute study and were compensated with a gift voucher. The study was approved by the local University's Research Ethics Committee.

\subsection{Study Set-up and Procedure}\label{sec:study:procedure}

\begin{figure}[t]
  \begin{subfigure}[t]{0.5\linewidth}
        \centering
        \includegraphics[width=\linewidth]{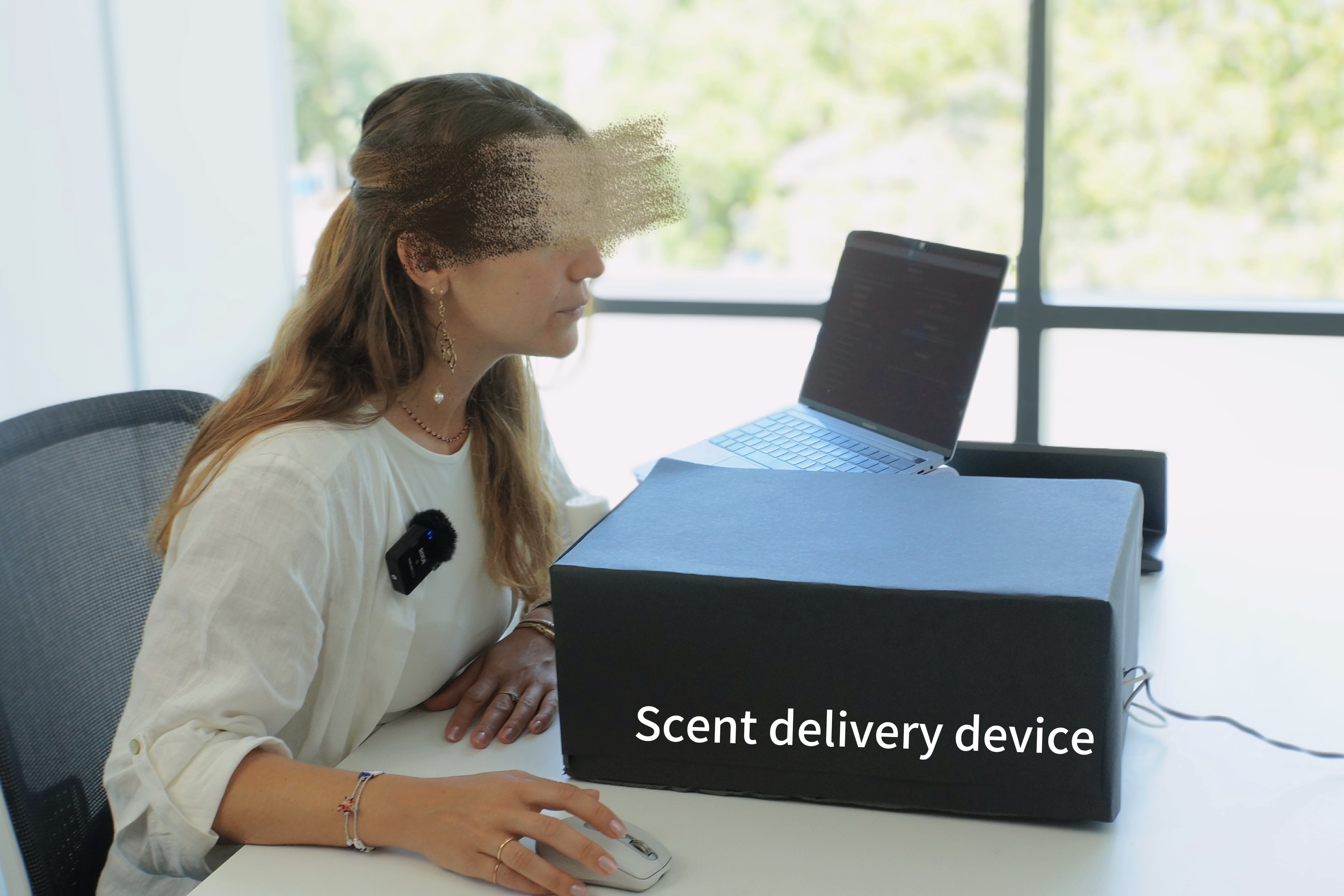}
        \caption{User study setup}
        \label{fig:user}
    \end{subfigure}%
    ~ 
    \begin{subfigure}[t]{0.5\linewidth}
        \centering
        \includegraphics[width=\linewidth]{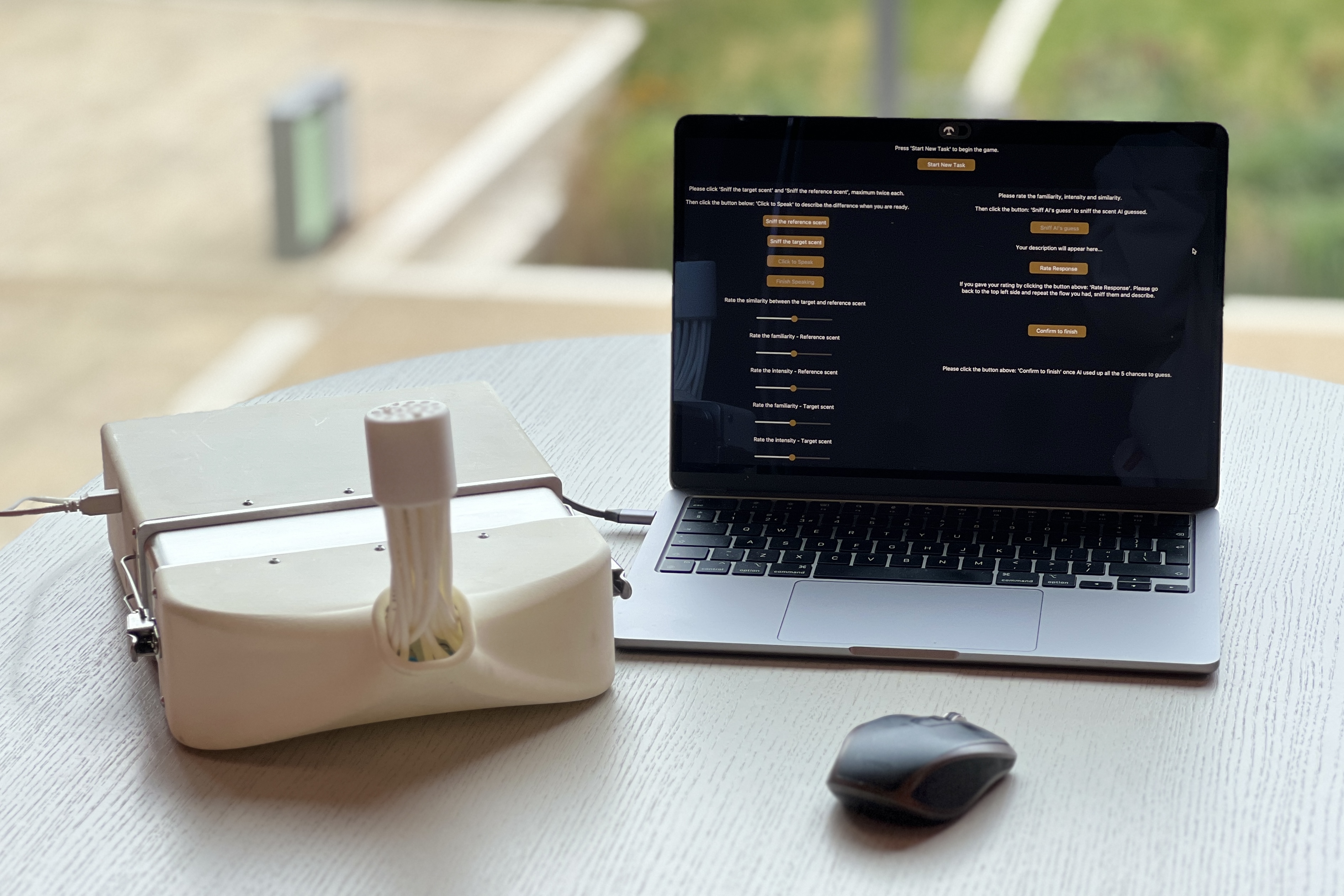}
        \caption{A close view of the device and user interface.}
        \label{fig:ui}
    \end{subfigure}
  \caption{Study Setups (a) A participant controls the AI guessing system, and sniffs the scent delivered from the device inside a noise isolation box. (b) A close view of the scent delivery device without the noise isolation box (left) and the user interface (right).}
  % \Description{Study Setups (a) A participant controls the AI guessing system, and sniffs the scent delivered from the device inside a noise isolation box. (b) A close view of the scent delivery device without the noise isolation box (left) and the user interface (right).}
  \label{fig:setup} 
\end{figure}

\begin{figure}[t]
  \centering
  \includegraphics[width=\linewidth]{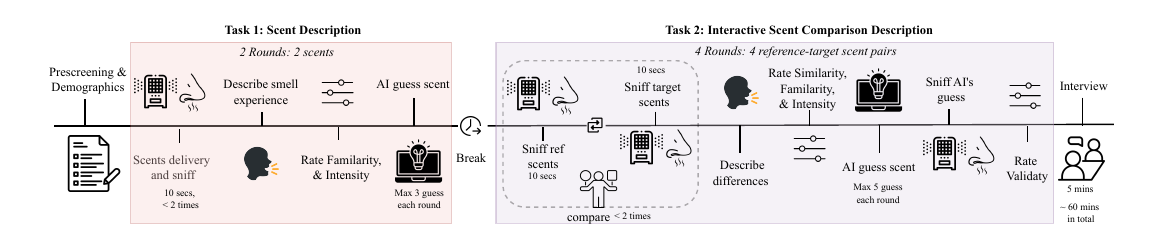}
\caption{User study procedure in four stages: 1. Prescreening \& Demographics, 2. Task 1: Participants sniff scents and describe them, followed by AI predictions, 3. Task 2: Participants compare scents, describe differences, and AI makes further guesses, 4. Concluding Interview.}
% \Description{User study procedure in four stages: 1. Prescreening \& Demographics, 2. Task 1: Participants sniff scents and describe them, followed by AI predictions, 3. Task 2: Participants compare scents, describe differences, and AI makes further guesses, 4. Concluding Interview.}
\label{fig:procedure} 
\end{figure}

We hosted the Sniff AI system on a local machine (13in, MacBook Pro Intel Core i7), allowing participants to complete tasks autonomously. Participants were provided with on-screen instructions to initiate the study, diffuse the scent, sniff it, and then describe it to the system. Participants also provided their subjective judgments via questions and rating options available on the interface, as shown in \Cref{fig:ui}. The detailed procedure of the study is illustrated in \Cref{fig:procedure}.

% Participants followed on-screen instructions to dispose scents and control their delivery via a novel scent-delivery device.  \Cref{fig:method} provides the work of the two tasks involved in the study. We now discuss study mechanics - before starting the tasks, participants first completed a prescreening and demographics questionnaire to confirm eligibility and gather background information. Participants were introduced with the scent deliver device - by clicking interface button, it can deliver scent to sniff, and they were asked to keep around 20 cm away from the device nozzle, the scent will be delivered for 10 seconds long. 

\paragraph{Pre-screening and Setup} Before joining the study, participants completed a pre-screening survey to ensure they had no olfactory or speaking impairments that could affect their participation. Upon arrival, they filled out a demographics questionnaire to collect essential background information. Participants were then introduced to the scent delivery device, as shown in \Cref{fig:user}, and instructed to maintain a distance of approximately 20 cm from the output nozzle. Each scent was delivered for ten seconds upon activation via an interface button to ensure standardized exposure throughout the study. To ensure a balanced evaluation, we limited the number of guesses the AI system could make in each round to three for Task 1 and five for Task 2. This threshold was established based on internal pilot testing, aimed at balancing the AI's opportunities with maintaining participant engagement.

Participants were reminded that the task focused on investigating AI's ability to learn about human sensory descriptions, rather than the human ability to recognize scents. They were instructed to avoid naming the scent (e.g., lime) in their descriptions and instead focus on describing the characteristics of the scent they experienced (e.g., sweet, punchy), how it made them feel (e.g., pleasant, unpleasant), and any other sensory qualities they observed.

\paragraph{Task 1: Scent Description} Each participant completed two rounds of Task 1. In each round, participants sniffed a target scent \(i_{tar}\) delivered via the novel scent-delivery device and described their olfactory experiences aloud to the Sniff AI system. The AI processed these descriptions and made an informed guess, presenting the results audibly and visually through an interface. Participants rated the scent on intensity and familiarity for the first sniff. If the AI correctly identified the scent, the round was considered complete. If the AI's guess was incorrect, participants had the opportunity to refine their descriptions, allowing the AI system to make up to three additional attempts per round. Subsequently, a different scent was introduced to continue with the next round.

\paragraph{Task 2: Interactive Scent Comparison} Participants completed four rounds of the Interactive Scent Comparison Task, each involving a new pair of reference and target scents. After sniffing both scents, participants described their differences and rated each scent on familiarity, intensity, and similarity. The AI then attempted to identify the target scent by asking, "Is this your target scent?" and delivering its guessed scent through the device. If the AI's prediction was correct, the round was completed, with the validity and similarity scores automatically recorded as 10. If the guess was incorrect, the mistakenly guessed scent became the new reference for subsequent descriptions and ratings. Both similarity and validity scores are key metrics for evaluating human-AI perceptual alignment. Definitions of these metrics are detailed in \Cref{sec:study:eval}.

\paragraph{Interview} After completing both tasks, participants were invited to a follow-up interview. 
% These interviews aimed to explore their perceptions of the AI’s performance, assess how well it aligned with human scent perception, and identify potential areas for improvement.

\subsection{Evaluating AI Sniff Performance}\label{sec:study:eval}
The study used a mixed-methods approach with a repeated-measures design to evaluate the performance of the LLM encoder. To ensure comprehensive coverage across scents and their families and minimize comparison biases, Task 1 involved randomly assigning two target scents to each participant while counterbalancing the total number of scents tested. Task 2 utilised a Latin Square arrangement with additional intra-group sequencing. This structured approach helped effectively control confounding variables through deliberate pairing and repeated-measures design, ensuring that each scent family, as well as each scent, was equally represented in the total number of trials. 
% We focused on the LLM encoder's ability to accurately identify scents across both tasks, examining the validity of its guesses and their similarity to the actual target scents in Task 2, based on user ratings. 
To measure the degree of perceptual alignment between humans and the AI model, we used the following evaluation metrics.

\paragraph{AI Success Rate} The success rate was calculated by dividing the number of correct predictions by the total number of rounds in each task. This metric aims to quantify the AI's effectiveness in accurately identifying scents.

\paragraph{Validity Scores} In Task 2, each AI guess receives a validity score, measuring how well the guess aligns with the human description by human expectation. A higher score indicates the AI accurately understood the user's input and chose a relevant scent. The validity of each AI guess is rated on an 11-point scale, ranging from 0 ("Completely Invalid") to 10 ("Completely Valid"). Correct predictions automatically receive a score of 10, while participants rate incorrect predictions.
% We analyzed the correlation between validity scores and success rates by examining average scores across guesses.

\paragraph{Similarity Scores} In Task 2, each pair of reference scent and target scent (including the initial reference) receives a similarity score where participants evaluate the perceived similarity between them. This score reflects how closely the AI’s guess matches the actual scent's characteristics. The similarity of each pair is rated on an 11-point scale, ("Not Similar at All") to 10 ("Completely Similar"). Similar to validity, correct predictions automatically receive a score of 10, and participants rate incorrect predictions. Our AI system learns from cumulative descriptions to predict a target scent (see \Cref{sec:method:prediction}), with each similarity score representing a different scent pair. By analyzing these scores when the AI guesses incorrectly, we can assess whether its guesses are progressively closer to the target scent, even when not perfectly accurate.

\paragraph{Familiarity and Intensity Scores} In scent-related studies, familiarity and intensity are commonly used metrics. Familiarity refers to how well participants recognize a scent, while intensity measures the perceived strength of the scent. We hypothesize that a higher familiarity with a scent may lead to a higher success rate in identifying or describing it, as people tend to describe things more accurately when they are familiar with them \cite{fiske1979person}.
% Both metrics provide essential insights into the perceptual experience of scents, helping to understand how recognisable and impactful the scents are to individuals.

\paragraph{Qualitative Insights from Semi-Structured Interviews} 
After completing the tasks, participants took part in semi-structured interviews to provide qualitative feedback on their experience with the AI system. These interviews explored their perceptions of the AI’s performance, its alignment with human scent perception, areas for improvement, and their vision for future human-AI scent interactions. The questions asked during the interviews are provided in the Supplementary Materials.

\section{Study Results}
We first present the overall performance of the Sniff AI system in both tasks. In Task 1 (Scent Description), the AI system made 213 guesses in 80 rounds, averaging 2.66 guesses per round (SD = 0.67). Task 2 (Interactive Scent Comparison Description), it made 648 guesses in 160 rounds, averaging 4.05 guesses per round (SD = 1.47).
Next, we provide a detailed analysis of scent-specific performances, categorizing results into scent families (e.g., Fresh) as shown in \Cref{tab:scent}; exploring the validity scores for the guessed scents and similarity between the reference and target scents as rated by human participants. Finally, we present qualitative insights from the interviews capturing their subjective experiences and feedback on the AI guesses. 
% Given the limited alignment observed, we conducted further exploratory analyses to investigate potential reasons behind this observed misalignment in \Cref{sec:results:ai}.

\subsection{Success Rates for Sniff AI Tasks}\label{sec:results:overacc} 

\subsubsection{Overall performance}\label{sec:results:overacc:over} 

The AI's overall performance was measured by its success rates across the two tasks. The success rate was calculated as the number of correct predictions divided by the total rounds for each task. 
In Task 1 (Scent Description), The AI system correctly identified the target scent in 22 out of 80 rounds, achieving a success rate of 27.50\% (SD = 0.29). The correct round count was predicted with 1 guess (9 times), 2 guesses (9), 3 guesses (4). For Task 2 (Interactive Scent Comparison), the overall success rate was 37.50\% (SD = 0.23) for 60 correct predictions out of 160 rounds. The correct round count was predicted with 1 guess (19 times), 2 guesses (14), 3 guesses (12), 4 guesses (10), and 5 guesses (5). A two-proportion z-test suggested that the performance increase in Task 2 was not statistically significant (z = 1.54, p = 0.0618).

We then categorized this rate per scent family as detailed in \Cref{tab:results_byscet}. For Task 1, the success rates per scent family—Fresh, Floral, Oriental, and Woody—were 40.00\%, 30.00\%, 25.00\% and 15.00\%, respectively. To determine the statistical significance of these differences, we conducted a Chi-Squared test, resulting in $\chi^2 = 8.43$ ($p = 0.75$). Task 2 exhibited rates of 45.00\%, 42.5\%, 40.0\% and 22.5\% for the same categories, with a Chi-Squared test resulting in $\chi^2 = 16$ ($p = 0.38$). No statistical signification was found between families.
% Fresh scents achieved the highest success rates in both tasks, and the statistically significant Chi-Squared values indicate notable differences in performance across the scent families in both tasks.

\subsubsection{Scent-specific Performance}\label{sec:results:byscent} 

\begin{figure}[t]
  \begin{subfigure}[t]{0.5\linewidth}
        \centering
        \includegraphics[width=\linewidth]{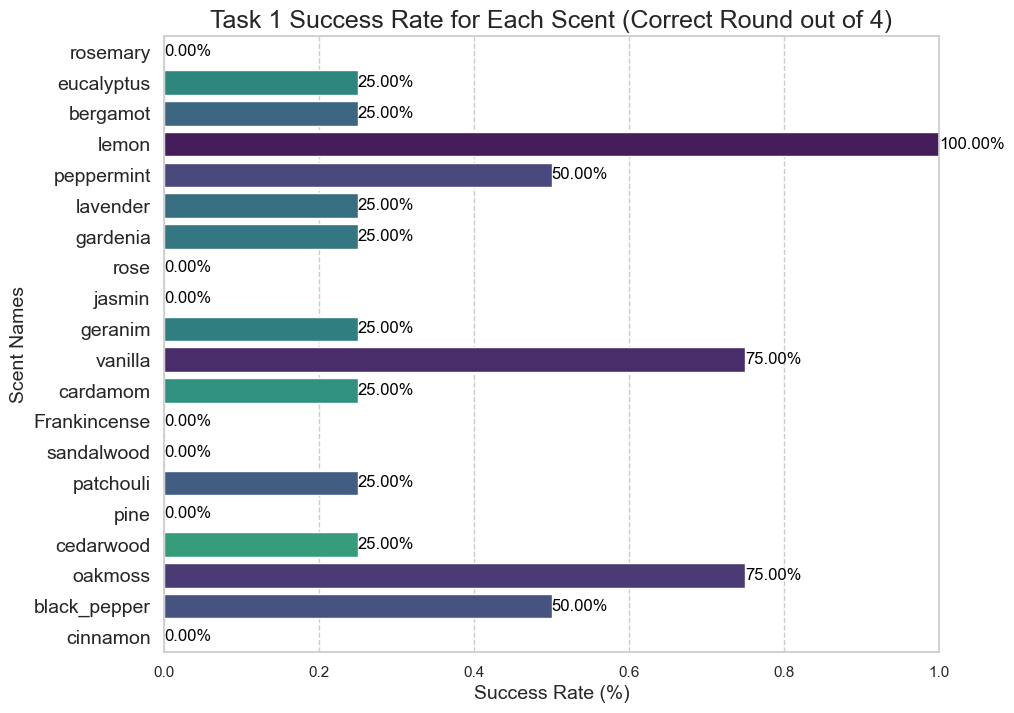}
        \caption{Success rates for each scent in Task 1.}
        \label{fig:acc_1}
    \end{subfigure}%
    ~ 
    \begin{subfigure}[t]{0.48\linewidth}
        \centering
        \includegraphics[width=\linewidth]{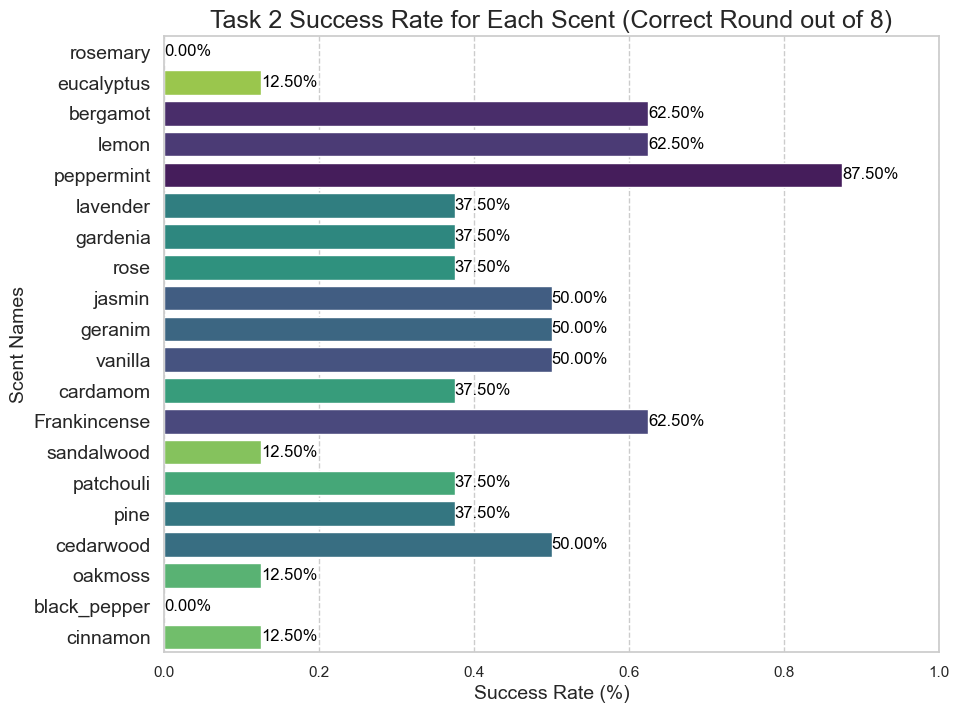}
        \caption{Success rates for each scent in Task 2.}
        \label{fig:acc_2}
    \end{subfigure}
  \caption{Success rate for each individual of the 20 scents used in the user study.}
  % \Description{Success rate for each individual of the 20 scents used in the user study, also detailed in table 2}
  \label{fig:acc} 
\end{figure}

Delving into the specific scents themselves, we present the descriptive summarised in \Cref{fig:acc_1}, \Cref{fig:acc_2} as well as \Cref{tab:results_byscet}. As shown, there were vast differences in success rates. For example, in Task 1, Lemon (ID 4) was identified in all rounds, whereas scents such as Rose (ID 8), Jasmin (ID 9), and Rosemary (ID 1) were never identified. Whereas in Task 2, Peppermint (ID 5) was identified for most of the rounds and achieved the highest success rate at 87.5\%, while neither rosemary (ID 1) nor black pepper (ID 19) were never identified with the lowest at 0\%.

\subsubsection{Confusion Matrices} 

\begin{figure}[t]
  \begin{subfigure}{0.5\linewidth}
        \centering
        \includegraphics[width=\linewidth]{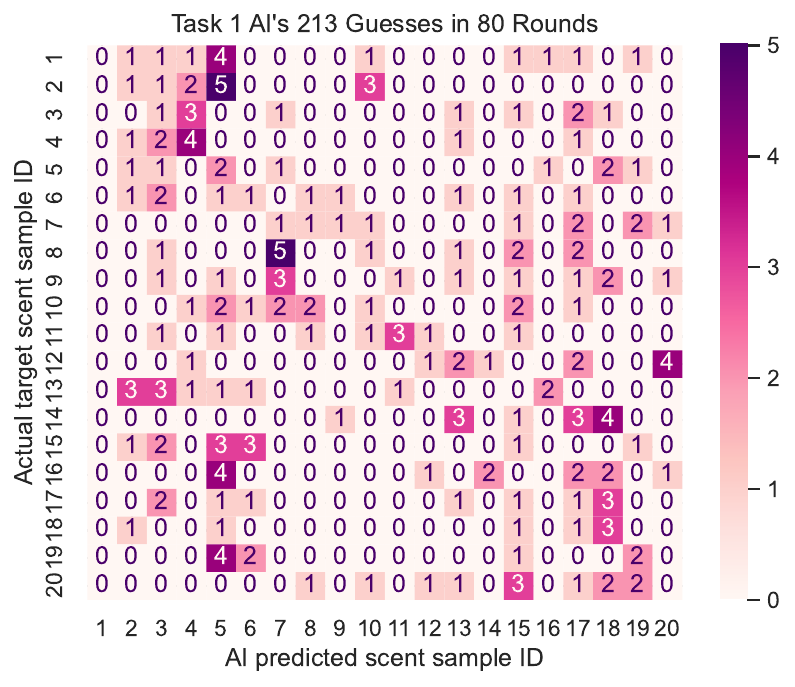}
        \caption{Confusion matrix for overall scent prediction in Task 1}
        \label{fig:cm1_s}
    \end{subfigure}%
    ~ 
    \begin{subfigure}{0.5\linewidth}
        \centering
        \includegraphics[width=\linewidth]{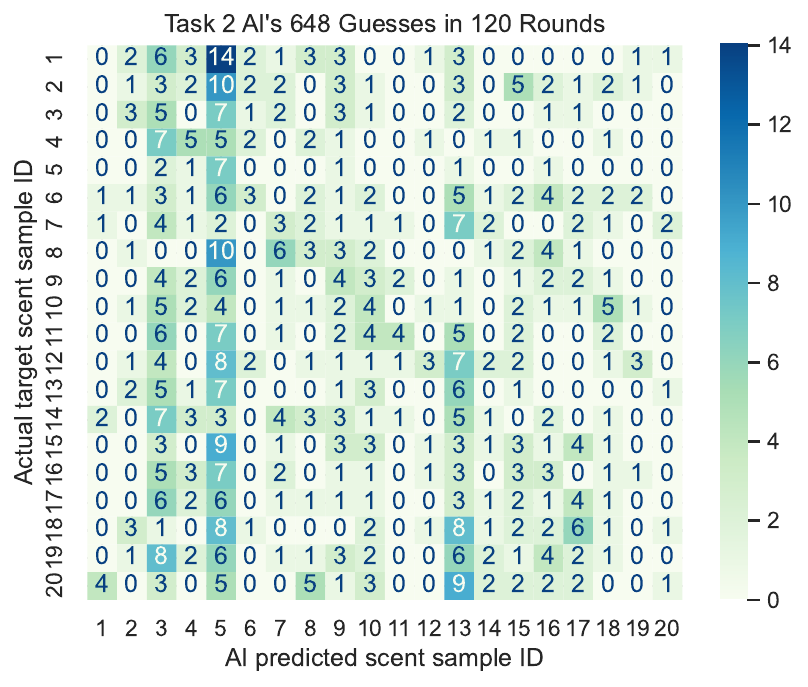}
        \caption{Confusion matrix for overall scent prediction in Task 2}
        \label{fig:cm2_s}
    \end{subfigure}
  \caption{Confusion matrices for overall scent prediction in Task 1 and Task 2.}
  % \Description{Two confusion matrices for overall scent prediction and actual target pairs in Task 1 and Task 2.}
  \label{fig:cm_scent} 
\end{figure}

Next, we created confusion matrices to understand when the AI was incorrect, which scent it would most commonly guess, to understand how the AI was misaligned. The confusion matrices that compare the AI's predictions (guesses) to the actual target scents for both tasks are displayed in \Cref{fig:cm1_s} for Task 1 and \Cref{fig:cm2_s} for Task 2. The horizontal axes represent the actual scent sample IDs, and the vertical axes show the AI's predicted scent IDs (guessed IDs). Each cell in the matrix represents how many times the AI predicted a specific scent ID (row) for a given actual scent (column). Darker cells indicate a higher frequency of predictions. The diagonal cells (from top left to bottom right) represent correct predictions where the actual scent matches the predicted scent (correct predictions). If human-AI alignment was strong, it would be expected a less total guess and to see a dark diagonal line from the top left to the bottom right of the graph. 

As the scent IDs are grouped by family, a slight deviation from this pattern would suggest that incorrect guesses were more commonly with same-family scents, compared to scents from a different family. From these graphs, for example, we see a bias for the AI model to incorrectly guess peppermint (ID 5) for the actual targets of rosemary (ID 1), Eucalyptus (ID 2), Pine (ID 16) and Black pepper (ID 19), where are all fresh or trigeminal scents. Similarly, the actual scent of Gardenia (ID 7) was most commonly guessed as Rose (ID 8).

Additionally, some columns show higher totals, indicating a bias in the AI's predictions towards certain scents. For example, in Task 1, the AI never guessed Rosemary (ID 1) but guessed Peppermint (ID 5) 30 times. In Task 2, the AI predicted Peppermint (ID 5) 137 times, while it made very few guesses for Cinnamon (ID 20), with only 5 predictions.

\subsection{Familiarity Scores}\label{sec:results:fam}
\begin{table}[t]
\centering
\caption{Results by Scent: the success rate (acc), familiarity score for the target scent across both tasks, and validity score for the target scent in Task 2.  * is overall success rate grouped by family}
\label{tab:results_byscet}
\adjustbox{max width=\textwidth}{%
\begin{tabular}{@{}cllccc|cccc@{}}
\toprule
\multicolumn{1}{l}{\multirow{2}{*}{Families}} &
  \multirow{2}{*}{ID} &
  \multicolumn{1}{l}{\multirow{2}{*}{Scent}} &
  \multicolumn{3}{c|}{Task 1} &
  \multicolumn{4}{c}{Task 2} \\
\multicolumn{1}{l}{} &
   &
  \multicolumn{1}{l}{} &
  ACC \% * &
  acc \% &
  Familirity &
  ACC \% * &
  acc \% &
  Familirity &
  Validity \\ \midrule
       & 1  & rosemary     &       & 0.00   & 6.00 ± 2.94 &       & 0.00  & 6.50 ± 2.73 & 4.68 ± 2.63 \\
       & 2  & eucalyptus   &       & 25.00  & 7.00 ± 3.37 &       & 12.50 & 5.13 ± 3.04 & 6.00 ± 2.64 \\
Fresh  & 3  & bergamot     & 40.00 & 25.00  & 5.75 ± 2.21 & 45.00 & 62.50 & 6.25 ± 1.83 & 6.54 ± 2.79 \\
       & 4  & lemon        &       & 100.00 & 8.50 ± 1.73 &       & 62.50 & 5.00 ± 3.07 & 6.50 ± 2.79 \\
       & 5  & peppermint   &       & 50.00  & 7.00 ± 3.56 &       & 87.50 & 5.25 ± 3.01 & 8.08 ± 2.30 \\ \midrule
       & 6  & lavender     &       & 25.00  & 9.00 ± 1.41 &       & 37.50 & 4.63 ± 2.97 & 5.92 ± 2.77 \\
       & 7  & gardenia     &       & 25.00  & 5.25 ± 1.71 &       & 37.50 & 6.25 ± 1.91 & 5.70 ± 2.70 \\
Floral & 8  & rose         & 15.00 & 0.00   & 4.75 ± 2.63 & 42.50 & 37.50 & 4.25 ± 2.76 & 4.03 ± 3.42 \\
       & 9  & jasmin       &       & 0.00   & 4.50 ± 3.11 &       & 50.00 & 4.13 ± 2.42 & 4.38 ± 3.02 \\
       & 10 & geranium     &       & 25.00  & 6.25 ± 2.87 &       & 50.00 & 5.75 ± 3.01 & 5.72 ± 2.92 \\ \midrule
       & 11 & vanilla      &       & 75.00  & 6.75 ± 2.87 &       & 50.00 & 5.88 ± 1.73 & 4.45 ± 3.04 \\
       & 12 & cardamom     &       & 25.00  & 6.25 ± 1.71 &       & 37.50 & 6.00 ± 2.73 & 6.41 ± 2.81 \\
Oriental &
  13 &
  frankincense &
  25.00 &
  0.00 &
  5.25 ± 1.71 &
  40.00 &
  62.50 &
  4.25 ± 2.96 &
  5.93 ± 3.30 \\
       & 14 & sandalwood   &       & 0.00   & 5.75 ± 3.03 &       & 12.50 & 6.25 ± 2.25 & 4.64 ± 2.39 \\
       & 15 & patchouli    &       & 25.00  & 5.25 ± 2.06 &       & 37.50 & 6.00 ± 2.39 & 4.42 ± 3.16 \\ \midrule
       & 16 & pine         &       & 0.00   & 6.25 ± 1.71 &       & 37.50 & 5.38 ± 2.50 & 5.83 ± 2.65 \\
       & 17 & cedarwood    &       & 25.00  & 6.50 ± 3.00 &       & 50.00 & 5.00 ± 2.62 & 5.77 ± 2.46 \\
Woody  & 18 & oakmoss      & 30.00 & 75.00  & 5.50 ± 2.38 & 22.50 & 12.50 & 5.25 ± 2.55 & 5.51 ± 2.77 \\
       & 19 & black pepper &       & 50.00  & 4.25 ± 3.30 &       & 0.00  & 5.25 ± 2.25 & 4.00 ± 2.97 \\
       & 20 & cinnamon     &       & 0.00   & 6.00 ± 3.16 &       & 12.50 & 4.25 ± 2.25 & 4.13 ± 2.70 \\ \bottomrule
\end{tabular}}
\end{table}
To determine whether the success rate is related to participants' familiarity with the scent, we analyzed the relationship between familiarity scores and success rates for the target scents in both tasks using Pearson correlation. In Task 1, the Pearson correlation showed a moderate but non-significant positive correlation (Pearson $r = 0.4004$, $n = 20$, $p = 0.0802$). Task 2 exhibited a weak, non-significant negative correlation (Pearson $r = -0.1577$, $n = 20$, $p = 0.5068$), suggesting no reliable linear relationship.

\subsection{Validity Scores for Interactive Scent Comparison (Task 2) and its Relationship to the AI's Performance}

\begin{figure}[t]
  \begin{subfigure}[t]{0.49\linewidth}
        \centering
        \includegraphics[width=\linewidth]{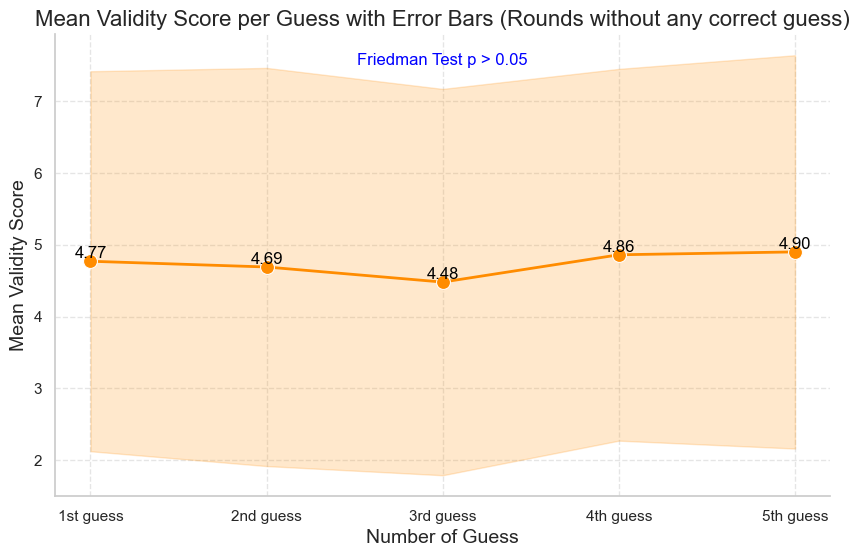}
        \caption{Mean validity scores per guess with error bars representing standard deviations (Where AI failed to guess correctly).}
        \label{fig:val_by_acc}
    \end{subfigure}%
    ~ 
    \begin{subfigure}[t]{0.49\linewidth}
        \centering
        \includegraphics[width=\linewidth]{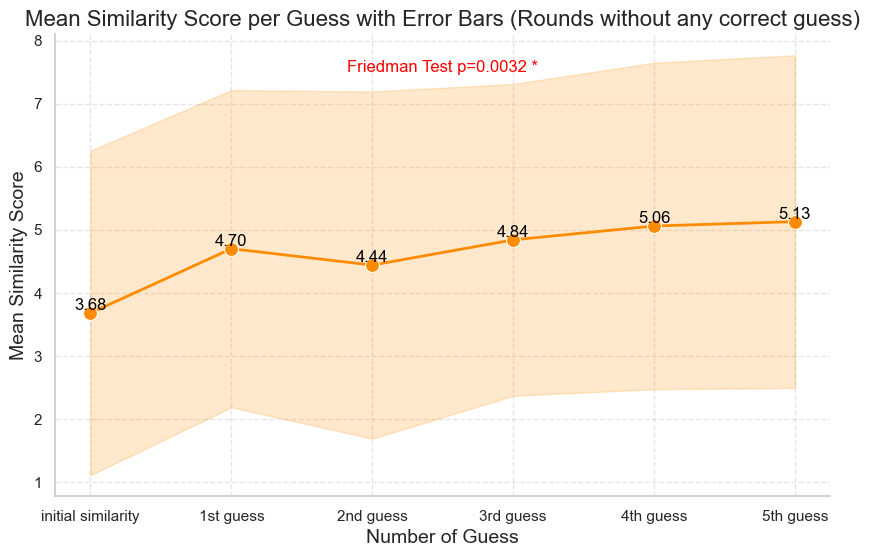}
        \caption{Mean similarity scores from initial part to per guess with error bars representing standard deviations (Where AI failed to guess correctly).}
        % \label{fig:sim_by_attempt}
    \end{subfigure}
  \caption{Mean validity (a) and similarity (b) scores plot with error bar and Friedman test (Where AI failed to guess correctly).}
  % \Description{Mean validity (a) and similarity (b) scores plot with error bar and Friedman test (Where AI failed to guess correctly).}
  \label{fig:by_attempt} 
\end{figure}

The validity score assesses how well the AI’s interpretations (guessed scent's characteristics) align with the participants' expectations in relation to their given descriptors. In Task 2, the overall validity score was 5.30 (SD = 2.98), which falls between "neither valid nor invalid" and "marginally valid". A one-sample t-test revealed that this mean score was significantly different from the neutral midpoint of 5 at a minimal effect size (\( t(647) = 2.55, p = 0.011, d = 0.10 \)), suggesting overall users rated the AI's judgments as marginally valid. We then exclude the successful round and explore the validity score during those rounds that AI failed in any guess; all scores fall around  "marginally invalid" as illustrated in \Cref{fig:val_by_acc}. Due to the non-normality in the data (Shapiro-Wilk test, all \textit{p} < 0.05), we used the Friedman test, which has non-significant differences across guesses  (\(p > 0.05\)).

We further calculated the relationships between the average validity score and success rate across all scents to understand how subjective perceptions of validity correlate with objective AI performance. This helps determine if participants' perceptions of the AI’s interpretations agree with its actual effectiveness. The results showed a moderate, significant positive correlation (Pearson $r$ = 0.64, $n$= 20, \(p = 0.0026\)). This indicates a significant positive relationship between these variables, supporting that subjective perceptions positively relate to AI effectiveness.

\subsection{Similarity Scores for Interactive Scent Comparison (Task 2)}\label{sec:valid_similar}
The similarity score measures the perceptual similarity between two scents as judged by participants. We focused on the similarities across guesses and explored whether the AI model improved in providing a more closely matched scent to the user's description. The average similarity score for the AI's guesses is 5.33 (SD = 2.92, excluding the similarity between initial target and reference), which falls between "neither similar nor dissimilar" and "marginally similar". A one-sample t-test revealed that this mean score was significantly different from the neutral midpoint of 5 at a minimal effect size, (\( t( 647 ) = 2.88, p = 0.004 \), d = 0.11), suggesting overall users rated the guessed scents as slightly more similar than neutral.

To analyze if there was an increase in similarity as the task progressed, we compared these ratings for where the AI did not correctly predict the scent, as successful rounds may have fewer than five guesses. \Cref{fig:by_attempt} shows the mean similarity scores from "initial similarity" to the "5th guess", with error bars representing standard deviations. 

Due to the non-normality in the data (Shapiro-Wilk test, all \textit{p} < 0.05), we used the Friedman test, which indicated significant differences across guesses (\(p = 0.0032\)). Follow-up Wilcoxon signed-rank tests with False Discovery Rate (FDR) corrections revealed that similarity scores significantly improve from the initial similarity (mean = 3.68, SD = 2.57, "marginally dissimilar") to the 1st guess (mean = 4.70, SD  = 2.52, "neither similar nor dissimilar", \textit{p} = 0.0051, r = -0.3084). However, scores did not improve at any of the subsequent guesses (all $p$ > 0.05, $r$ < -0.13); suggesting that users did not feel the AI improved in providing them a more similar scent after the initial guess. 

% Despite this, a significant improvement was observed when comparing the initial similarity score to the 5th guess (\( p = 0.0022 \), $r = -0.3814 $), indicating a notable overall improvement in perceived similarity across attempts. The negative effect size suggests that the perceived similarity decreased moderately to substantially from the initial guess to the fifth guess. 

\subsection{Interview results}\label{sec:results:interview}
\begin{table}
\centering
\caption{Interview reflections, including the addressed questions }
\label{tab:codebook}
\adjustbox{max width=\textwidth}{%
\begin{tabular}{@{}llcl@{}}
\toprule
Addressed Question (Description) &
  Subthemes &
  Frequency &
  Examples of Quotes \\ \midrule
 &
  High &
  10 &
  \begin{tabular}[c]{@{}l@{}}"The AI has been guessing well, in some cases guessing closer and correct" (P03);\\ "I think maybe 70\% of the time it moved in the right direction." (P06);\\ "very impressed by AI's ability to exactly understand what I was saying. " (P14)\end{tabular} \\ \cmidrule(l){2-4} 
\begin{tabular}[c]{@{}l@{}}To what extent does users perceive \\ the AI alignment to be accurate?\end{tabular} &
  Medium &
  20 &
  \begin{tabular}[c]{@{}l@{}}"many times the AI is guessing towards the correct direction, but the extent of \\ its guessing is not well-controlled." (P05);\\ "It listened to the words I was saying and on one occasion it was correct."(P22); \\ " The AI was sometimes good and sometimes not." (P34)\end{tabular} \\ \cmidrule(l){2-4} 
 &
  Low &
  10 &
  \begin{tabular}[c]{@{}l@{}}"Inaccurate roughly speaking, it was converging towards, mostly diverged from \\ the description." (P01); \\ "If I speak generally, the AI cannot guess accurately." (P24);\end{tabular} \\ \midrule
 &
  Emotions &
  5 &
  \begin{tabular}[c]{@{}l@{}}"when I said pleasant.. I confused it..I had to sort of avoid using emotive \\ language and try and stick to the like hardcore descriptors" (P10)\\ "I think the AI does not have a clear concept of how humans think that it is \\ a good smell and attacking smell, or whether it is a bad smell." (P12)\end{tabular} \\ \cmidrule(l){2-4} 
\begin{tabular}[c]{@{}l@{}}What specific instances where\\the AI met, or failed to meet, \\ users expectations in capturing \\the nuances of different scents?\end{tabular} &
  \begin{tabular}[c]{@{}l@{}}Personal \\ experiences\end{tabular} &
  6 &
  \begin{tabular}[c]{@{}l@{}}"like bakery and coffee shop, and it didn’t really get those references" (P07)\\ "I said it would be used in cooking, and then it would give me a scent that would \\ never be used in cooking." (P25)\\ "we describe to a person which may share our experience so they may understand \\ better because we live in the same environment and have to share" (P40)\end{tabular} \\ \cmidrule(l){2-4} 
 &
  Perceptions &
  13 &
  \begin{tabular}[c]{@{}l@{}}“...the AI is quite obsessed with minty and non-minty flavour...” (P08)\\ "one flavor is more citric than the other and also more creamy \\ that's sweet so they are would often doesn't get it right." (P17)\\ " I think AI has kind of done a great job in trying to kind of identify ..\\ the freshness or maybe how minty it smells." (P35)\end{tabular} \\ \midrule
 &
  \begin{tabular}[c]{@{}l@{}}Difficulty \\ verbalising \\ the smell\end{tabular} &
  28 &
  \begin{tabular}[c]{@{}l@{}}"...too hard for humans to describe the amount of actions that needs to be taken \\ precisely to describe it." (P20); \\ "...I struggled a little bit with trying to describe certain smells..." (P27)\end{tabular} \\ \cmidrule(l){2-4} 
\begin{tabular}[c]{@{}l@{}}What are users' reflections\\on interacting with AI\\in scent experience?\end{tabular} &
  \begin{tabular}[c]{@{}l@{}}Interpreting \\ about AI's \\ behaviour\end{tabular} &
  27 &
  \begin{tabular}[c]{@{}l@{}}"either approximately or distance from the target scents"(P01) - \textit{(Diverging)}\\ “shocked AI is obsessed with minty and non-minty flavour” (P08) - \textit{(Biased word)}\\ "keeps coming back to the wrong answer that it gave previously, ..my guess is that\\it was trapped in some local minimum" (P18) - \textit{(Repetitive Jumping)}\\ "Even though I didn't always get it right, I could see almost the logic between the\\ steps it was taking." (P39); - \textit{(Miscellaneous)}\end{tabular} \\ \cmidrule(l){2-4} 
 &
  \begin{tabular}[c]{@{}l@{}}Novel Human-AI\\ interaction\end{tabular} &
  15 &
  \begin{tabular}[c]{@{}l@{}}"it's quite interesting and it's actually a bit surprising that the AI actually can get \\ sometimes a bit closer after each trial. So, I think that's very surprising to me." (P17)\\ "interesting to interact with AI, try to use clear phrases that it can understand" (P25)\\ "very novel, interesting and is my first time to attend a smelling experiment." (P31)\end{tabular} \\ \midrule
 &
  Entertainment &
  34 &
  \begin{tabular}[c]{@{}l@{}}"Selecting perfume for customers." (P05); \\ "scent may also be used in the XR areas... make the scent experience more natural \\ and more likely to the real world." (P28)\end{tabular} \\ \cmidrule(l){2-4} 
\begin{tabular}[c]{@{}l@{}}What potential applications do \\users envisage for human-AI\\olfactory interactions in the future?\end{tabular} &
  Healthcare &
  8 &
  \begin{tabular}[c]{@{}l@{}}"...potentially in helping identify allergies in future." (P16); \\ "Perhaps people who have a problem with their smell, it might help them \\ to appreciate smells." (P22)\end{tabular} \\ \cmidrule(l){2-4} 
 &
  \begin{tabular}[c]{@{}l@{}}Safety \\ Regulation\end{tabular} &
  11 &
  \begin{tabular}[c]{@{}l@{}}"...like we have a sniff dogs at airport...AI can do that as well." (P13); \\ " ...detecting maybe dangers...like explosives or things like that... fire...sewage \\ problems (P35)\end{tabular} \\ \bottomrule
\end{tabular}}
\end{table}
We summarized the results of our semi-structured interviews in \Cref{tab:codebook}, using thematic analysis with double-blind coding by two co-authors. Four main themes were discovered in a bottom-up fashion presented in \Cref{tab:codebook}, under the Addressed Question section. Each theme consists of three subthemes: (1) \textit{Perceived degree of human-AI alignment}, which categorised participant's views into High, Medium, and Low; (2) \textit{Unaddressed factors in AI scent selection}, including Emotion, Personal Experiences, and Perception; (3) \textit{User reflections on AI scent interaction}, with subthemes of Difficulty in Verbalizing Experiences, Observed AI Behaviour; (4)\textit{Future prospects of Human-AI olfactory interaction}. Specifically, to better understand AI’s behaviour, we identified 4 subthemes: Words, Deviations from Descriptions, Repetitive Jumping, and Miscellaneous (comments on AI’s working mechanism). To ensure consistency, we counted the frequency of each theme once per participant, even if they raised multiple points related to the same theme.

We observed varying degrees of alignment between participants' perceptions and the AI's guesses. Over half of the participants (n=30) reported lower than expected. Many (n=18) emphasised the importance of emotional attachment and personal sensory experiences in understanding scents - areas where AI struggled. This was particularly evident as the AI failed to capture nuances in scent categories and intensities, relying mostly on concrete terms to describe physical attributes. Consequently, AI guesses often deviated from participants' descriptions (n=21). The AI also showed a pattern of repetitive and alternating guesses, shifting between nearly correct scents and unrelated ones (n=6). Additionally, nearly half of the participants (n = 19) reported difficulty verbalising their olfactory experiences due to limited vocabulary, lack of prior experience describing scents, and the transient nature of the experience. Despite these challenges, 14 participants found the interaction novel and engaging. Participants also anticipated potential future applications for AI-driven scent technologies in entertainment (n=34), healthcare (n=8), and safety regulation (n=11), although the AI’s current limitations in processing subjective and culturally contextual scent descriptions were clear.

\subsection{Replacing Human Participants with GenAI Models}\label{sec:results:ai}

Given the limited alignment observed in Task 1, we conducted further exploratory experiments. Specifically, we designed a simple experiment using language generative models to perform Task 1 (Scent Description) instead of human participants -- describing its interpretation of each selected scent. We consider the current most powerful LLM families $f_{LLM}$: OpenAI GPT \cite{achiam2023gpt},  Google Gemini \cite{team2023gemini} and Anthropic Glaude \cite{anthropic2024introducing}, and compare the following models via their official APIs:

\begin{itemize}
    \item OpenAI GPT-4 and GPT-4o \cite{achiam2023gpt}: \texttt{gpt-4-turbo-2024-04-09} and \texttt{gpt-4o-2024-08-06}
    % "gpt-3.5-turbo"
    \item Google Gemini 1.0 and 1.5 \cite{team2023gemini}: \texttt{gemini-1.0-pro} and \texttt{gemini-1.5-pro}
    \item Anthropic Claude 3 and 3.5 \cite{anthropic2024introducing}: \texttt{claude-3-opus-20240229} and \\
    \texttt{claude-3-5-sonnet-20240620}
\end{itemize}

We prompt these LLMs to describe how a scent smells without mentioning the name, aiming to reflect the description of an average person, each result in $f_{LLM}(x_i)$. \Cref{equ:encode} would then generate a set of vectors: $\mathcal{V}_{LLM} = \{f_{encoder}(f_{LLM}(x_1)), ..., f_{encoder}(f_{LLM}(x_{20})) \}$ where each value in $\mathcal{V}_{LLM}$ represents an embedding vector generated by encoding the response returned by an LLM. The prompts and experimental settings are detailed in supplementary material.

\begin{figure}[t]
  \begin{subfigure}[t]{0.49\linewidth}
        \centering
        \includegraphics[width=\linewidth]{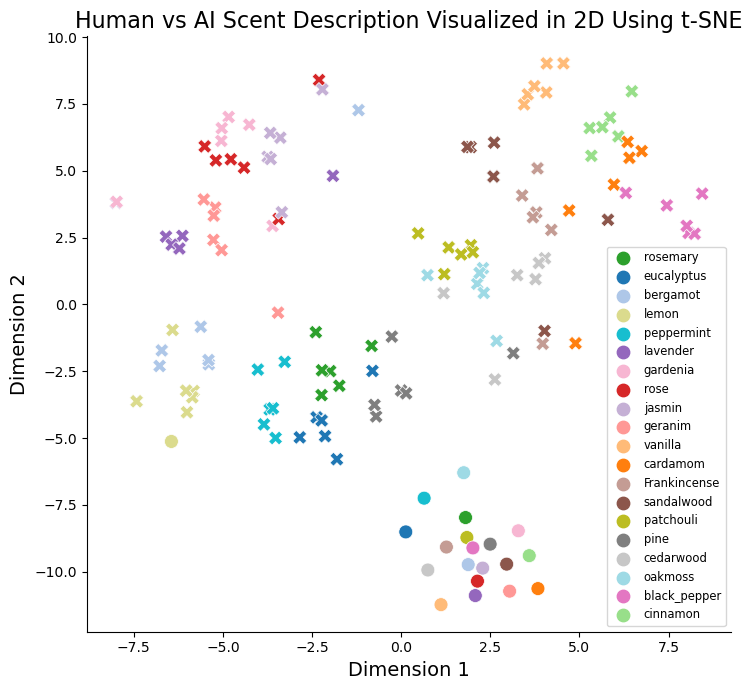}
        \caption{Human (dots) and LLMs (crosses) scent description t-SNE.}
        \label{fig:ait}
    \end{subfigure}%
    ~ 
    \begin{subfigure}[t]{0.44\linewidth}
        \centering
        \includegraphics[width=\linewidth]{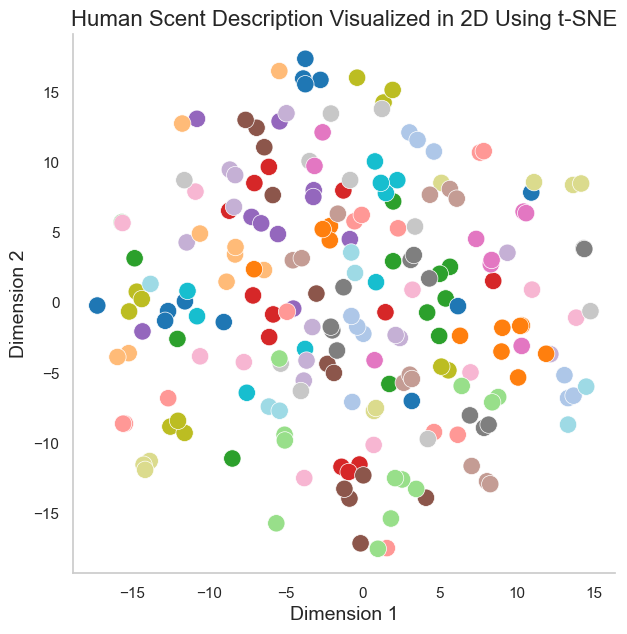}
        \caption{Human smell experience description across 20 scent t-SNE.}
        % \label{fig:humant}
    \end{subfigure}
  \caption{2D t-SNE for human and AI scent descriptions.}
  % \Description{t-SNE for human and AI scent descriptions.}
  \label{fig:tsne} 
\end{figure}

To compare the descriptors generated by LLMs  $\mathcal{V}_{LLM}$ with humans $\mathcal{V}_{human}$. We employ the t-distributed Stochastic Neighbor Embedding (t-SNE) algorithm, a commonly used explainable AI tool, to visually compare their embeddings in 2D. It is an unsupervised dimensionality reduction that transforms high-dimensional Euclidean distances between data points into conditional probabilities that reflect their similarities in low dimension, in our case, 2D.

% \subsubsection{Human vs LLMs Expressions}

We first calculate the centroid of the vectors for human descriptions of each scent. These centroids capture the average semantic space of human perceptions for each scent. We then compared and visualized these centroids with $\mathcal{V}_{LLM}$ in the same embedding space using t-SNE as illustrated in \Cref{fig:tsne}. This allows us to observe the clustering and dispersion patterns, comparing how closely AI-generated descriptions resemble human perceptions. 
Most GenAI-generated points are far from human points, suggesting that the AI's scent representations are not well aligned with human sensory experiences. However, for the scent "lemon," where the AI achieved a 100\% success rate, the points for GenAI and human descriptions are close, becoming the only point with a high alignment.

% Most of the AI-generated descriptions are far away from humans, whereas only the lemon is close.

\begin{figure}[t]
  \begin{subfigure}[t]{0.49\linewidth}
        \centering
        \includegraphics[width=\linewidth]{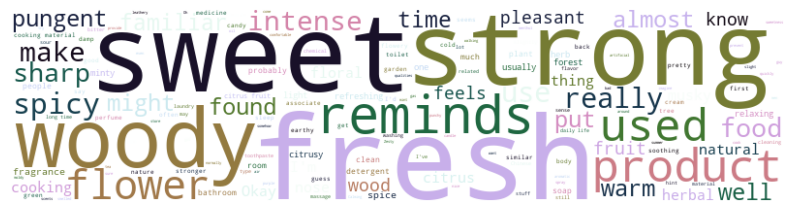}
        \caption{Wordcloud for human}
        \label{fig:wc_human}
    \end{subfigure}%
    ~ 
    \begin{subfigure}[t]{0.49\linewidth}
        \centering
        \includegraphics[width=\linewidth]{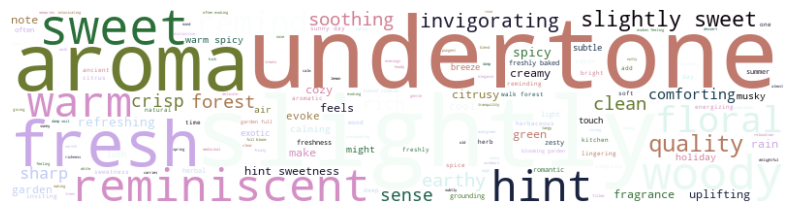}
        \caption{Wordcloud for AI}
        \label{fig:wc_ai}
    \end{subfigure}
  \caption{Wordclouds for both human and AI performing Task 1}
  % \Description{Wordclouds for both human and AI performing Task 1}
  \label{fig:wcloud} 
\end{figure}

Additionally, we explore the linguistic differences between GenAI and human descriptions by analyzing term frequency. We first excluded non-substantive words (e.g. "it", "this") and study-specific terms (e.g. "feel", "smells") using Python NLTK. We then used WordCloud for visualisation. This approach emphasizes on the most significant content words used in descriptions, providing insight into the focus and variability of language used as depicted in \Cref{fig:wcloud}. The word cloud for humans prominently features words like "sweet", "fresh", "strong", "woody" and "reminds". AI highlights terms such as "undertone", "aroma", "slightly", "reminiscent" and "hint".

\section{Discussion}
The primary objective of this work was to assess how well LLMs align with human in smell experiences. We conducted a user study where participants sniffed and described scents, and an LLM-based embedding model integrated into an AI system guessed the scents in real-time. Task 1 focused on accessing LLM encoder's ability to match a specific scent based on human-provided description during the study, in other words, its general semantic understanding of scent. In Task 2, we evaluate the LLM encoder's ability to understand and represent the relationships between different scents.
% In Task 1, users sniffed and described a target scent up to three times, with the AI model guessing the scent based on user descriptions from a catalogue of 20 possibilities. In Task 2, users described the difference between a target and a reference scent, with the AI again guessing the target scent based on these descriptions. 
The AI system demonstrated moderate success in identifying scents, with an overall success rate of $27.50\%$ in Task 1 and rate of 37.50\% in Task 2. Our results indicate that LLMs can, to some extent, represent scent semantics within their embedding spaces, though this alignment is limited and biased toward certain scents. There is some alignment and promising potential, but significant challenges remain for LLM to fully understand human smell experiences. Below, we discuss reasons for this limited alignment and how the future would improve performance.

\subsection{AI's Performance in Understanding Scents}\label{sec:dis:understand} 
A basic measurement for AI's understanding of scent is through our Scent Description Task (Task 1) as discussed in \Cref{sec:method:map}. Our quantitative findings presented in \Cref{sec:results:overacc} show that there is limited alignment in the LLM encoder model to comprehend smell experiences, with only a 27.50\% success rate in Scent Description (Task 1). Also, most scents received only one correct guess (8 instances) or none at all (7 instances); suggesting human-AI alignment varies depending on the specific scent. Notably, the encoder frequently confused eucalyptus (ID2) with peppermint (ID5) and gardenia (ID7) with rose (ID8), misclassifying each pair five times; we further discuss this divergence in \Cref{sec:dis:bias}.
% Despite both pairs belonging to the same scent families, Fresh and Floral families respectively, it is still indicating that the encoder has only a moderate ability to identify scents based on human descriptions.

This limited performance could be due to limitations in the LLM's ability, and also the challenges humans face in describing scents. Scents lack a standardized language and are often linked to personal experiences and memories (e.g., events, times, and people) as also suggested from our interview \Cref{tab:codebook} \cite{schifferstein2005capturing, strauch2022olfactory, lledo2005information}. People tend to describe scents by referencing their sources, using tangible objects to help others understand the smell \cite{drake2003flavor,zarzo2008relevant}. In our study, participants were instructed not to name the scent but to describe the things that came to their mind, focusing on scent characteristics. Half of the participants (n=19) mentioned their difficulty verbalising the smell experiences as shown in \Cref{tab:codebook}.
They employ direct, sensory-focused terminology like "fresh", "strong", and "woody" and often relate to "remind" of their personal experience (see \Cref{fig:wc_human}). These descriptions are straightforward and resonate with everyday experiences. This may challenge LLMs to match the scent, as these personal experiences may be their unseen scenarios. For example, the AI did not associate "sandalwood" with descriptions such as \textit{"an expensive candle or home incense" (P7), but it did successfully link "lavender" to }" a teddy bear sleep product" (P38) as shown in \Cref{fig:teaser}. Additionally, human descriptors focus on the immediate sensory impact: for example, describing the rose as \textit{"a little bit sweet and maybe purple"} (P23, AI misrecognise it as geranium (ID7)) or noting that \textit{"it smells like some kind of flowers. The smell is quite light, not heavy at all, and it makes me feel very relaxed and comfortable"} (P29, AI misrecognise it as gardenia (ID10).

To further investigate the abovementioned phenomenon, we have AI tackled Task 1 similarly to a human participant, as detailed in \Cref{sec:results:ai}. As shown in \Cref{fig:wcloud}, LLMs tend to utilize more abstract, expert terminology such as "undertone," which may feel detached from common everyday usage and occasionally lack intuitive sense. We found that LLMs provide a richer narrative that is sometimes distant from typical human descriptions. For instance, GPT-4 describes a rose as \textit{"sweet and floral, like a blooming garden full of delicate petals, carrying a soft, romantic fragrance with a hint of a nutty undertone"} (\texttt{gpt-4o}). Similarly, Claude-3 proposes \textit{"a delightful floral aroma that is both sweet and subtly spicy, evoking the essence of a lush garden in full bloom, with a warm, nutty undertone that adds depth"} (\texttt{claude-3-opus}). 
This difference is important to consider, as AI-generated descriptions might not always align with how people naturally talk about scents, and there exists a challenge to map personal experiences with scents (also reflected from our interviews \Cref{tab:codebook}), all these factors may affect the effectiveness in everyday human-AI communication. This phenomenon is likely caused by the fact that these large models are pre-trained on more professional, formal, or technical datasets. We will further discuss this in the limitation section.

Additionally, the t-SNE visualization (\Cref{fig:tsne}) clearly demonstrates a separation between AI-generated and human-given scent descriptions. Interestingly, lemon, identified 100\% in the study, is an exception where human descriptions cluster closely to those of the AI, suggesting alignment in human-AI descriptions for this scent. Despite this anomaly, the separation highlights a significant gap in how the AI and humans perceive and describe scents. This observation aligns with findings from Zhong et al. \cite{zhong2024feeling}, who explored AI's abilities in describing tactile sensations. Moreover, our results indicate that AI models show more variance in tasks related to scent than those involving tactile descriptions. This could be due to the models being trained on different data sources. Also, descriptions of scents vary more than tactile descriptions, leading to less consistent performance across various models.

\subsection{Limited Alignment in Smell Experiences}
The Interactive Scent Comparison Task (Task 2) evaluated the degree of alignment between the LLM's scent representations and human descriptions, focusing on the embedding model's dimensional structure, through comparative descriptions. Although the success rate of 37.50\% marks an improvement from Task 1, it still remains moderate. We observe that LLM's performance is improved across all scent families (Fresh, Floral, Oriental, and Woody), with the most notable gains in the Floral category. This improvement could be due to participants gaining experience in communicating about scents from Task 1. Additionally, the presence of a reference point in Task 2 seemed to aid participants in describing scent differences, as suggested by our interview data. Despite these insights, the specific reasons for the marked improvement in the Floral family are not fully understood and need further investigation.

% By verbalizing the differences between scents, we are interested in exploring whether the predicted scents were moving in the direction described by participants, for understanding the LLM's internal representation and relationships of scents. 
To address the challenges of articulating scent experiences as discussed in \Cref{sec:dis:understand}, we employed subjective assessments such as validity and similarity scores during the study. For instance, when describing the comparison between geranium and pine, a participant (P30) noted, "The target scent [geranium] is less spicy, more flowery, though not pleasantly so, and more akin to nature." Based on this description, the AI predicted the scent as cedarwood and, accordingly, diffused cedarwood. Although the AI's guess was incorrect, participants rated this prediction as significantly valid (8) but significantly dissimilar (2). In this case, participants found the prediction is valid in terms of interpret their description - the cedarwood about "less spicy, more flowery, not pleasantly, akin to nature", but not perceptual similar to the target. We still view this as an aligned case as AI gave a scent that largely matched human description.

Essentially, the validity score reflects the AI's success in interpreting human descriptions, while the similarity score indicates how well the AI matches the scent's perceptual profile. Overall, participant ratings for the AI’s interpretation of validity and similarity were slightly above neutral, leaning "marginally valid/similar". Although the AI occasionally aligned with participants' descriptions, its overall capability to match human scent perceptions remained limited as validity scores hovered around marginally invalid during unsuccessful rounds. We then focused on identifying the trend for perceptual alignment concerning similarity. Our system and task design enables the AI to refine its understanding through cumulative descriptions. We tracked changes in similarity scores over time, as shown in \Cref{fig:by_attempt}. 
Initially, there was a significant increase in similarity, followed by a decrease between the first and second guesses. The scores showed minimal iterative improvement but were not significantly different. This pattern could be attributed to our study's design, which balanced reference and target pairs through intra-group sequencing. Each participant encounters three pairs from different groups and one from the same group. Although the AI's guesses progressed to be more perceptually similar to the target, the increased guesses needed for a correct prediction (\Cref{sec:results:overacc:over}) suggests that accuracy is most likely on the first attempt. Subsequent attempts may not necessarily enhance prediction accuracy.  This pattern raises questions about the cumulative effectiveness and the LLM’s ability to understand context, especially considering the lack of significant changes in similarity after the initial guess. If the AI's first output is inaccurately directed, it becomes challenging to provide progressive and cumulative contrastive explanations that guide the AI toward the correct answer.

% Further insights into AI's limited scent perception regards capture the nuances regarding emotion, perception during interviews. Many participants noted that the AI often focuses on certain explicit characteristics, such as specific nouns or adjectives, while ignoring broader contextual meanings. For example, a participant (P25) mentioned, "When I said it's not a green smell, it's not a flowery smell, and then it gives me a flowery smell, it might be just because it's picking up on the keywords". 
% Moreover, discussions with researchers highlighted further issues in the dimensional structure of current LLMs embeddings. They note that LLMs sometimes align with the intended directions, they often fail to capture subtle nuances like the variance in scents. Several computer scientist pointed out flaws in the AI's optimization processes. For example, P17 mentioned that the AI sometimes gets trapped in a local minimum, leading to inaccurate responses. P20 observed that excessive step changes during optimization can cause the model to overshoot the global minimum, resulting in settling at a local minimum.
% These findings highlight challenges in how LLMs assign "weights" to different characteristics and their limited understanding of scent profiles, confirming the existing limitations in their embedding models.

More importantly, olfaction alignment is relatively low compared to other modalities such as vision \cite{lee2023visalign} or sound \cite{peter2023we}. Also, Marjieh et al. \cite{marjieh2023large} showed that GPT-4 could effectively interpret human sensory judgments (e.g., colour, sound, and taste) based on textual inputs. This divergent performance in olfaction may stem from the underrepresentation of smell experiences in AI training data and the focus of AI on linking olfactory descriptors to chemical structures in the domain \cite{keller2017predicting, lee2023principal}. As a result, there has been a surge in calls for datasets and labelling efforts aimed at empowering AI to "sniff" with olfactory descriptions, like the DREAM challenge \cite{keller2017predicting} and the Odeuropa project \cite{oderopa2020smell}.

\subsection{AI Exhibits a Bias toward Specific Scents and Particular Scent Characteristics}\label{sec:dis:bias}
The success rates of both tasks varied across the four scent families and individual scents, indicating the AI’s biases in these tasks. The LLM encoder excels with Fresh family scents but struggles with Floral family scents. The performance also varies with Woody family scents. When considering individual scents, the AI shows high accuracy with distinctive and common scents such as lemon and peppermint. This variability in performance not only suggests a potential bias but also points to gaps in the AI's ability to consistently process different scent categories. 

We first explore whether "bias" arises from AI's limited capability or originates from participants themselves. Previous studies suggest people tend to describe things more accurately when they are familiar with them \cite{fiske1979person}. However, we found no significant correlation between success rate and familiarity. In some cases, the results were even divergent; for example, Task 1 with the highest familiarity, Lavender (ID 6), had only a 25\% accuracy rate, while the task with the lowest familiarity, Black Pepper, achieved a 50\% accuracy rate (see \Cref{tab:results_byscet}).

We then investigate the confusion matrix in \Cref{fig:cm_scent} at where AI had difficulty distinguishing between scents. Both confusion matrices in \Cref{fig:cm_scent} present a bias in the perception alignment across scents, where we observed a scattering of bright spots rather than a concentrated diagonal line in those square matrices. Ideally, in an instance of perfect alignment between the AI and human judgments, the confusion matrix should exhibit its highest values only on the diagonal line running from the top left to the bottom right. Our confusion matrices suggest that while the AI is somewhat aligned with human judgments for some scent families, there is significant variability, especially in distinguishing certain scent families such as Floral and Woody. For example, the AI frequently confused peppermint (ID 5) with both rosemary (ID 1) and eucalyptus (ID 2). This confusion could stem from their similar cool scent profiles. Similarly, gardenia (ID 7) was often mistaken for rose (ID 8), likely due to their perceptual similarity within floral characteristics. It was also observed that the AI repeatedly guessed peppermint (ID 5) in both tasks, one participant described the AI as "obsessed with minty and non-minty" (P08). This may be attributed to peppermint's distinctive minty/fresh characteristics and its prevalence in training data. As a result, the AI often mislabels other scents as peppermint, indicating a bias toward more familiar or common descriptions.

Interview feedback from participants also highlighted an observed bias in the AI towards particular scents, especially those that are more uniquely identifiable, such as mint and lemon. Then participants included descriptors related to \textit{"minty"} or \textit{"citric"}; the AI often defaulted to predicting peppermint or lemon, even if these descriptors were used to indicate an absence of these qualities. For instance, participants noted that, \textit{"identify citrusy more accurately than other scents”}(P2), \textit{"whenever I mention citric it will give me the smell of some kind of lemon" (P17) and "AI is quite obsessed with minty and non-minty flavour"} (P08). This also confirms our finding in LLM's limited contextual understanding of scent profiles and relationships toward literal interpretations.
Conversely, the AI struggled significantly with scents like rosemary, which had a $0\%$ success rate in both identification tasks. This suggests that certain scents may lack the distinctive characteristics that the AI can readily identify or are underrepresented in the training data. Such discrepancies highlight the AI’s difficulty in recognizing less distinctive or less commonly trained scents.

\section{Limitations and future work}
This study provides insights into the AI's capabilities and current limitations in scent recognition. However, it is important to acknowledge several challenges that might affect the broader applicability of these findings. First, we have a limited scent sample size that is based on the Fragrance Wheel \cite{edwards2010fragrances}.
% while 40 participants, goes beyond typical HCI studies \cite{hornbaek2006current}, it is small in the context of AI and LLM research. 
Future work is needed to extend the sample selection and explore a larger diversity of scents. This could be done through combining in-person studies, as in our work, with online surveys, where more descriptions on peoples' smell experiences can be collected \cite{obrist2014opportunities}. 
% This could also help to enhance the scent samples.

Second, this study primarily recruited non-experts to reflect the AI's intended use by the general public, focusing on human-like rather than "textbook" language. This only represents the general population and not necessarily professionals, limiting the generalizability of the conclusions.
% However, participants often struggled to describe scents accurately due to limited vocabulary and the abstract nature of olfactory descriptions (see \Cref{sec:results:interview}). This linguistic limitation often led to imprecise and abstract scent descriptions that might not fully capture the intended essence of the scent experience. 
Future work could compare the same settings with both experts and non-experts to better understand how expertise influences AI's performance in scent recognition. 

Third, to enhance AI’s capabilities in processing scent-related language, it is crucial to enrich training data with a diverse array of descriptive terms and emotional expressions associated with scents. This aims to reduce biases and limitations evident in LLMs. Current research in computer science typically focuses on chemical descriptors, which are not readily applicable to everyday uses by the general public. Future studies could improve by integrating multimodal data inputs, and combining chemical scent analysis with human descriptions to create a more comprehensive AI-based scent prediction framework. 
Furthermore, employing Human-in-the-Loop strategies like Reinforcement Learning from Human Feedback (RLHF) \cite{ouyang2022training} can enable AI systems to adapt and improve their predictions continuously based on user feedback, gradually increasing their effectiveness and relevance. Last but not least, there is growing interest in olfactory experiences within the HCI community \cite{brooks2023third, becsevli2024smell, maggioni2020smell}. 

% Participants (n=15) expressed enthusiasm about the engaging nature of human-AI interaction involving scents. They described the experiment as \textit{"very fun"} and \textit{"interesting"}, and appreciated exploring scents in \textit{"novel"} ways (P31, P37). Based on participant feedback, there are potential applications for scent-focused AI models in various sectors, including entertainment, healthcare, and safety. For example, AI could help individuals choose their favourite perfumes or assist those with olfactory impairments in experiencing different scents (P18, P22). Additionally, similar to sniffer dogs at airports, AI could detect substances for security purposes (P13). 

%We hope that addressing these areas in future research can enhance AI's capabilities in smell experiences, making it a valuable asset across industries such as perfumery, food and beverage, environmental monitoring, and healthcare. These advancements would not only improve technological efficiency but also deepen our understanding of sensory interactions between humans and machines.

\section{Conclusion}
In this work, we explored human-AI perceptual alignment in smell experiences. We developed an AI system, which leverages an LLM embedding model, named Sniff AI. Sniff AI recognizes human voice input, predicts, and delivers scents in real-time. 
To investigate how well the LLM encoder aligns with human perception, we conducted a user study involving 40 participants who interacted with the Sniff AI system. 
During the study, participants sniffed various scents and described them, whilst the AI attempted to identify them based on their descriptions. Our findings indicate that while the LLM's embedding space captures scent-related semantics, it exhibits limited accuracy and a bias toward certain scents. 
Additionally, participants responded positively to their interactions with Sniff AI, describing the experience as \textit{"very fun"} and \textit{"interesting"}, and appreciating the novel ways of exploring scents. Feedback highlighted the potential uses of scent-focused AI across entertainment, healthcare, and security sectors. 
For instance, AI might assist in choosing perfumes, aid those with olfactory impairments, or even detect substances in security settings. 
These findings and insights highlight both the potential and challenges of AI in aligning with human sensory experiences related to smell. Existing AI models demonstrate capability in recognizing distinct scents from descriptions; however, they encounter significant limitations in processing nuanced or subjective scent descriptions. This reveals critical avenues for future research to enhance AI's understanding of and interaction with human olfactory experiences.

\bibliographystyle{unsrt}  
\bibliography{references}

\end{document}